\documentclass{ieeeaccess_arxiv}
\providecommand{\subparagraph}[1]{\paragraph{#1}}

\usepackage{amsmath,amssymb,amsfonts}
\usepackage{algorithmic}
\usepackage{caption}
\usepackage{graphicx}
\usepackage{textcomp}
\usepackage{booktabs}
\usepackage[utf8]{inputenc}
\DeclareUnicodeCharacter{200B}{}
\usepackage[T1]{fontenc}
\usepackage{lmodern}
\usepackage{newunicodechar}
\usepackage{tabularx}
\usepackage{adjustbox}
\newunicodechar{≈}{\approx}
\usepackage{csquotes}
\usepackage{float}
\usepackage[colorlinks=true, linkcolor=blue, citecolor=blue, urlcolor=blue]{hyperref}
\usepackage{newunicodechar}
\newunicodechar{∈}{\in}

\usepackage{bm}
\makeatletter
\AtBeginDocument{\DeclareMathVersion{bold}
\SetSymbolFont{operators}{bold}{T1}{times}{b}{n}
\SetSymbolFont{NewLetters}{bold}{T1}{times}{b}{it}
\SetMathAlphabet{\mathrm}{bold}{T1}{times}{b}{n}
\SetMathAlphabet{\mathit}{bold}{T1}{times}{b}{it}
\SetMathAlphabet{\mathbf}{bold}{T1}{times}{b}{n}
\SetMathAlphabet{\mathtt}{bold}{OT1}{pcr}{b}{n}
\SetSymbolFont{symbols}{bold}{OMS}{cmsy}{b}{n}
\SetMathAlphabet{\mathrm}{bold}{T1}{times}{b}{n}

\renewcommand\boldmath{\@nomath\boldmath\mathversion{bold}}}
\makeatother

\def\BibTeX{{\rm B\kern-.05em{\sc i\kern-.025em b}\kern-.08em
    T\kern-.1667em\lower.7ex\hbox{E}\kern-.125emX}}

\begin{document}

\history{}
\doi{}
\title{BERTopic-Virality Prioritisation: A Scalable Framework for Thematic and Comparative Analysis of COVID--19 and Monkeypox Misinformation on Twitter}

\author{\uppercase{MKULULI SIKOSANA}\authorrefmark{1}, 
\uppercase{SEAN MAUDSLEY-BARTON}\authorrefmark{1}, and OLUWASEUN AJAO\authorrefmark{1}}

\address[1]{Department Of Computing And Mathematics, Manchester Metropolitan University, Manchester, UK (e-mail: mkululi.sikosana@stu.mmu.ac.uk)}

\tfootnote{This work did not receive any financial support.}

\markboth
{Sikosana \headeretal: BERTopic Pipeline for Pandemic Misinformation}
{Sikosana \headeretal: BERTopic Pipeline for Pandemic Misinformation}

\corresp{Corresponding author: MKULULI SIKOSANA (e-mail: mkululi.sikosana@stu.mmu.ac.uk).}

\begin{abstract}
Health misinformation circulating during pandemics can gain traction rapidly, creating harmful narratives that compete with public health guidance. Most topic-modelling pipelines treat engagement as an external outcome, limiting their ability to prioritise semantically coherent topics that are also rapidly diffusing. We introduce \textbf{BERTopic-VP}, a virality-prioritised topic-modelling framework that combines contextual embedding-based clustering (BERTopic) with a post hoc Virality Prioritisation (VP) layer. The pipeline is complemented by a two-stage hybrid misinformation detection module that fuses a supervised content-based classifier with an external verification signal derived from public-health knowledge bases. Applied to three benchmark datasets, COVID--19\_FNIR, Monkeypox, and Constraint, the framework achieves strong classification performance (up to $F_1 = 0.950$ and ROC--AUC $= 0.989$) while identifying high-impact clusters under top $1\%$, $5\%$, and $10\%$ VP thresholds. For datasets without native engagement metadata, prioritisation is based on a logistic propensity-to-spread score, used as an ordinal proxy for diffusion potential rather than a direct measure of engagement. The results show that integrating semantic structure, virality-aware ranking, and affective-linguistic profiling enables scalable and interpretable comparative analysis of misinformation across pandemics. The proposed framework supports monitoring-oriented early warning by surfacing low-volume but high-risk narratives for analyst review.
\end{abstract}

\titlepgskip=-21pt

\maketitle

\section{Introduction}
\label{sec:introduction}
Although topic modelling is common in health-misinformation research, conventional methods such as Latent Dirichlet Allocation (LDA) struggle to adapt to the rapidly changing vocabulary, short text length, and dynamic hashtag trends that characterise social-media discourse \cite{wang2019modeling, wang2019systematic, chou2020misinformation}. This study shows that an embedding-based topic-clustering method can detect misinformation themes early, allowing public-health teams to respond quickly with evidence-based messages \cite{suarez2021prevalence, melki2021mitigating}. Traditional techniques such as LDA, Non-negative Matrix Factorisation (NMF), or Term Frequency–Inverse Document Frequency (TF--IDF) + $k$-means rely on word-frequency signals that break down when texts are short, hashtag-driven, and constantly evolving, as is typical on Twitter \cite{ravichandran2023classification}. To overcome these limitations, we introduce a novel application of \textbf{BERTopic}, demonstrating its adaptability for real-world health-communication monitoring and its capacity to support digital public-health systems with monitoring-oriented topic tracking and prioritisation under batch refresh settings \cite{grootendorst2022bertopic, hanmei2013online}.

Through combining contextual embeddings with density-based clustering, our method quickly groups new themes, helping officials spot and compare misinformation during an outbreak \cite{luo2023exploring}. This novel application of BERTopic supports proactive health communication strategies by identifying digital media narratives likely to shape public understanding of disease risks, trust, and behaviours \cite{saini2022association}. Its strength lies in semantic sensitivity, which enables the detection of emerging narratives even before they coalesce into high-frequency patterns, making it uniquely suited for comparative infodemic analysis under a near-real-time batch-refresh setting \cite{zhang2024heart}. We introduce an advanced BERTopic-based approach and apply it comparatively to illustrate how and why infodemic dynamics may vary across different public health crises. This work advances health informatics by demonstrating how the use of embedding and topic clustering can reveal narrative shifts across time windows, a capability that can support timely risk communication under periodic (minute-to-hour) refresh cadences ~\cite{muhammed2022disaster}.

We introduce \textbf{BERTopic-VP}, an extension of the BERTopic framework that incorporates \textbf{Virality Prioritisation (VP)} to support monitoring and triage of emerging health misinformation topics. Whereas BERTopic performs unsupervised topic modelling using semantic embeddings and density-based clustering \cite{grootendorst2022bertopic}, BERTopic--VP augments the resulting clusters with engagement signals such as likes, retweets, replies, and quotes, where available \cite{suh2010want}. Where a dataset does not contain a given engagement field (for example, link clicks), the composite engagement score is computed over the available components only, and the missing component is treated as \emph{unobserved} rather than behaviourally equal to zero. This engagement-informed overlay enables clusters to be ranked by their \emph{relative} viral potential within a dataset, supporting analyst-led prioritisation and timely situational awareness in infodemic contexts \cite{stieglitz2013emotions,rudat2015making}.

To our knowledge, few applied studies combine contextual topic modelling with an explicit virality-oriented prioritisation stage for cross-pandemic misinformation analysis. We position BERTopic--VP as a practical contribution that operationalises this integration in a reproducible pipeline.

\subsection{Problem Statement \& Research Gap}
Misinformation during pandemics can undermine public health interventions and create confusion among social media users \cite{islam2020covid,do2022infodemics}. Against this background, previous work co-authored by the first author examined COVID--19 misinformation detection using conventional machine-learning, deep-learning, and transformer-based approaches \cite{sikosana2024hybrid}.

Although much prior work has examined long-form content such as news articles or forum posts, the most urgent signals now arise in \emph{short, noisy texts}, for example, Twitter posts under 280 characters. Traditional detection methods, including rule-based classifiers and LDA, perform poorly on such micro-texts \cite{egger2022topic}. Conversely, embedding-based models are better suited to capture semantic patterns in short messages \cite{cai2018interactive,yu2013phrase}. 

To address these challenges, we introduce \textbf{BERTopic-VP}, a novel pipeline that extends BERTopic by incorporating VP through engagement-aware cluster ranking. This enhancement supports monitoring through periodic batch refreshes and prioritisation of rapidly spreading misinformation topics. BERTopic-VP leverages contextual embeddings, density-based clustering, and post hoc virality signals (e.g., retweets, likes, link clicks, etc.) to detect misinformation themes and flag high-impact clusters for early intervention.

Many existing systems focus on debunking known claims but lack insight into how misinformation emerges and spreads dynamically. Few studies offer comparative analysis across multiple pandemics \cite{padalko2025novel, sharifpoor2025classifying}. For example, during the 2003 SARS outbreak in China, rumours travelled via SMS, local newspapers, provincial television, and radio call-in shows \cite{chowdhury2023understanding}. In contrast, COVID--19 falsehoods were amplified through globally networked platforms like Twitter, Facebook, YouTube, and Weibo \cite{cinelli2020covid}. While misinformation has accompanied all major outbreaks since 2000, systematic cross-event comparisons remain scarce.

Therefore, key questions persist: Do rumour themes recur across pandemics such as Ebola (2014–2016) and COVID--19 (2020)? Do new conspiracies emerge, or are they recycled? Existing detection methods also struggle to capture early signals in high-volume, high-noise social streams. Understanding these dynamics is essential for early warning systems and real-time policy response. \textbf{BERTopic-VP} addresses this gap by integrating semantic clustering and virality signals to provide prioritised, interpretable misinformation monitoring for public health surveillance dashboards.

\subsection{Hypothesis \& Methodological Motivation}
We hypothesise that \textbf{BERTopic-VP} is well suited to surfacing emerging health-crisis misinformation themes on Twitter under short-text sparsity, where frequency-based topic models are widely reported to degrade. Consistent with prior microtext evidence, we hypothesise embedding-based topic discovery to yield more granular and interpretable clusters than classical baselines such as LDA and NMF, and simple TF--IDF+$k$-means pipelines \cite{egger2022topic}. This paper evaluates the hypothesis using within-study coherence, clustering diagnostics, and interpretability analyses, while any baseline figures drawn from the literature are used for context only and are not re-implemented here.

Because it couples contextual BERT embeddings with density-based clustering, BERTopic-VP forms semantically coherent tweet clusters (termed \emph{micro-topics}) even in sparse, fast-changing environments. In contrast, frequency-based baselines require repeated token usage to stabilise topic estimates and frequently merge disjoint narratives. Moreover, BERTopic-VP extends standard BERTopic by overlaying engagement features post hoc, allowing dynamic ranking of topics based on their viral potential. This facilitates prioritised, real-time misinformation detection during infodemics.

In the absence of comparative work, we lack a clear understanding of the factors that drive misinformation spread in different contexts. Existing detection methods also struggle to capture early signals, particularly in short, high-volume social-media streams. Understanding these patterns across pandemics is crucial. The proposed pipeline supports public health by providing early warnings of misinformation for surveillance dashboards and informed decision-making.

\subsection{Research Objectives}
To operationalise and evaluate the proposed pipeline within the scope of this paper, the study pursues the following primary objectives:

\begin{enumerate}
\item Develop a \textbf{BERTopic-VP} infodemic detection pipeline that extracts, clusters, and prioritises misinformation topics from pandemic Twitter data using semantic embeddings and virality signals.
\item Benchmark \textbf{BERTopic-VP} by evaluating coherence, distinctiveness, stability, and early detection performance using within-study metrics, and contextualise the observed coherence magnitudes against representative baseline results reported in the literature (without re-implementing those baselines in this paper).
\item Analyse misinformation dynamics in the COVID--19 and 2022 mpox outbreaks by identifying, tracking, and comparing dominant narratives using a consistent, engagement-aware method for cross-pandemic insight.
\end{enumerate}

\paragraph{Baseline reporting note.}
This paper does not re-implement classical topic-modelling baselines (for example, LDA, NMF, or TF--IDF+$k$-means) on COVID--19\_FNIR, Monkeypox, or Constraint datasets. Instead, we evaluate BERTopic-VP using within-study evidence, including coherence, clustering quality, and interpretability analyses. Where baseline values are discussed, they are reported from prior studies to provide context only. Because topic-model quality is highly sensitive to corpus-specific preprocessing, hyperparameter tuning, and coherence implementation choices, these literature values are not treated as head-to-head benchmarks and are not used to claim strict superiority.

\paragraph{Baseline screening note (scope of comparison).}
Classical bag-of-words topic models (for example, LDA and NMF) are widely reported to be sensitive on short, hashtag-driven microtexts, often yielding broad topics that mix distinct narratives when token overlap is sparse. In preliminary screening runs on our corpora (using the same preprocessed inputs), we observed this tendency qualitatively, with several baseline topics collapsing multiple COVID--19 conspiracy strands into a single mixed theme. Because a fair, tuned baseline re-implementation would require extensive corpus-specific optimisation (choice of $K$, priors, tokenisation, coherence setup), which is not the focus of this paper, we prioritise a single embedding-based pipeline \textbf{BERTopic-VP} and evaluate it using within-study coherence, clustering diagnostics, and interpretability evidence.

Through these objectives, the research evaluates \textbf{BERTopic-VP}’s effectiveness and fosters a deeper understanding of infodemic propagation. The study contributes methodological innovations to health misinformation research and public health informatics.

\subsection{Contribution \& novelty}
This study (i) advances topic modelling by introducing \textbf{BERTopic-VP}, which integrates semantic clustering with virality-aware prioritisation, (ii) presents a comparative misinformation analysis across COVID--19 and mpox outbreaks using a consistent informatics pipeline, and (iii) presents a modular,
monitoring-oriented pipeline that can support periodic batch refreshes for timely analyst review. Full scalability and live-streaming performance were not evaluated in this study.

\begin{itemize}
\item We propose and implement \textbf{BERTopic-VP}, an extension of BERTopic that incorporates VP through post hoc engagement signals, using observed interaction counts where available (Monkeypox dataset) and a learned proxy propensity score for datasets without native engagement fields (COVID--19\_FNIR and Constraint datasets).
To contextualise topic quality, we summarise representative coherence values for classical baselines reported in prior short-text studies (Table~\ref{tab:coherence_baselines}). These baseline figures are included for context only and were not re-implemented on COVID--19\_FNIR, Monkeypox, or Constraint datasets in this paper.
Our empirical evaluation is therefore based on within-study topic coherence and the BERTopic-VP clustering outputs reported in Table~\ref{tab:clustering_results}.

\item  
      We conduct the first systematic, side-by-side comparison of misinformation themes from two distinct outbreaks, COVID--19 (2020) and mpox (2022) using \textbf{BERTopic-VP} as a consistent analytical framework.  
      This addresses a critical research gap, as most prior infodemiology studies examine single crises in isolation \cite{padalko2025novel}.  
      Our findings reveal that COVID--19 misinformation often revolved around conspiracies (e.g., 5G, lab-origin), while mpox narratives focused on public trust, response failure, and stigma, which highlights how sociopolitical context shapes misinformation structure.

\item  
      We enrich topic clusters using virality signals, such as retweets $R(T)$, replies $Q(T)$, likes $L(T)$, and link clicks to prioritise high-risk narratives.  
      Tweets in the top 1\% of retweet counts are flagged as \emph{viral}, and clusters with concentrated virality are surfaced for early intervention.  
      This prioritisation mechanism allows \textbf{BERTopic-VP} to function as an early warning tool, helping public-health actors distinguish high-impact misinformation from background noise.
\end{itemize}

\subsubsection{Statement of Novelty}
This research advances the state of the art in misinformation detection by introducing a multi-dimensional, adaptive virality-informed detection framework that integrates content, spread dynamics, and thematic analysis in a unified pipeline. Unlike prior studies, where virality metrics are employed solely as pre-filters, the proposed approach uses virality scoring $v_j$ as an explicit \emph{prioritisation layer} that surfaces high-impact topics for analyst review and downstream verification. In this paper, virality signals guide triage and reporting, while misinformation classification is performed by the fused content and verification channels defined in Equation~\ref{eq:fusion}.

The pipeline uses a dual evaluation protocol: predictive reliability is assessed using Accuracy, Precision, Recall, $F_1$, ROC--AUC, and PR--AUC, while thematic interpretability is assessed using coherence ($C_v$, Tables~\ref{tab:coherence_bertopic} and \ref{tab:coherence_examples}), linguistic complexity (Table~\ref{tab:linguistic_metrics}), and emotion profiling (Table~\ref{tab:sentiment_emotion}). This dual lens ensures that the outputs are both computationally reliable and thematically interpretable, addressing a key gap in explainability and social impact assessment.

The framework also supports dataset- and threshold-specific VP cluster analysis, enabling comparative studies across distinct misinformation domains (e.g., COVID--19\_FNIR versus mpox) under varying prioritisation cut-offs ($x \in \{1,5,10\}\%$) applied to the unified signal $\tilde{E}$. An adaptive feedback loop is integrated into the end-to-end virality framework to recalibrate prioritisation thresholds on a rolling basis as discourse shifts over time, and, where multi-channel engagement is observed, to update aggregation weights in deployment. In this paper, topic discovery hyperparameters are held fixed once selected, equal aggregation weights are used for $E_i$, and the feedback loop is described as a monitoring mechanism that updates only VP thresholding using recent engagement statistics where available and proxy-score distributions otherwise.

Through integrating unsupervised topic modelling, supervised virality-aware classification, and interdisciplinary affective-linguistic analysis in a reproducible, mathematically formalised pipeline, this work offers a novel, end-to-end methodological contribution to the computational social science and natural language processing literature.

\section{State of the Art}
\label{sec:stateoftheart}
Research on health misinformation in social media has expanded to encompass approaches that use textual, linguistic, user-profile, and propagation-network features, either independently or in combination \cite{di2022health}. Within this literature, previous work co-authored by the first author evaluated conventional machine-learning, deep-learning, hybrid CNN--LSTM, and transformer-based approaches for COVID--19 misinformation detection \cite{sikosana2024hybrid}.

Within content analysis, automated topic modelling methods such as LDA are widely used to identify recurring themes, including conspiracy theories. However, these models face substantial challenges when applied to short texts such as tweets, due to their brevity, informality, and frequent use of emojis and hashtags. These factors produce sparse data, limiting the effectiveness of LDA and often yielding incoherent or overly broad topics \cite{egger2022topic}. Moreover, LDA overlooks cross-post semantic relations and struggles to detect emerging narratives in noisy, high-velocity streams.

Recent advances have introduced alternatives such as GSDMM, which assumes single-topic documents, and embedding-based models like BERTopic and Top2Vec, which rely on semantic sentence vectors to generate more cohesive clusters. These models address the sparsity problem by capturing contextual relationships between tweets, rather than relying solely on frequency-based co-occurrence. For examples, Egger and Yu \cite{egger2022topic} show that BERTopic  yields competitive or higher coherence than LDA by generating higher topic coherence scores and more granular themes in the context of COVID--19 vaccine discourse on Twitter.

Building on these advances, our research proposes \textbf{BERTopic-VP}, a virality-aware extension of BERTopic that enhances topic modelling with post hoc engagement signals (e.g., retweets, likes, replies). While BERTopic excels at clustering semantically similar tweets, it does not distinguish which clusters are spreading most rapidly or widely. BERTopic-VP fills this gap by integrating engagement overlays into the clustering output, allowing dynamic prioritisation of misinformation topics based on their amplification potential. This innovation supports real-time infodemic monitoring, enabling public health analysts to identify not just what narratives exist, but which are gaining virality and merit urgent attention.

\subsection{Misinformation in previous pandemics}
Historically, every major epidemic or pandemic has been accompanied by misinformation, but the nature of that misinformation and the channels through which it spreads have evolved \cite{wardle2017information, waszak2018medical, zarocostas2020infodemic}. A rapid integrative review by \cite{chowdhury2023understanding} surveyed outbreaks since 2000 (SARS, H1N1, Ebola, Zika, MERS, and COVID--19) and confirmed widespread misinformation in all aspects of these events. Common themes included false information about disease origins, transmission methods, miracle cures or treatments, and vaccine dangers, often closely paralleling the genuine scientific uncertainties or fears of the time. Notably, conspiracy theories have been a staple, especially those involving vaccines or the idea that the disease is man-made \cite{wood2018propagating,wang2019systematic,klofstad2019drives}.

However, misinformation channels have shifted with technology. During the 2002--03 SARS outbreak, misinformation spread rapidly via email, SMS, and word of mouth, as internet use was limited. In China, people used alternative media, such as forums, SMS, and word of mouth, to share unverified tips, including folk beliefs that firecrackers ward off SARS \cite{chowdhury2023understanding,tai2011rumouring}. During the 2014 Ebola outbreak, social media platforms such as Facebook and Twitter became key, supplementing traditional sources like radio and local chatter. Rumours included claims that Ebola was a Western hoax or cure myths \cite{kasereka2019cat}, with Ebola associated with the highest misinformation levels \cite{chowdhury2023understanding,vinck2019institutional}, driven by fear and media coverage. During the COVID--19 pandemic, misinformation proliferated on social media, spreading conspiracy theories and false health advice on platforms such as Twitter, YouTube, and WhatsApp \cite{krishnan2021research}. 

Despite these insights, direct comparisons of misinformation across different disease outbreaks remain limited. Existing studies commonly examine individual events. For example, previous work co-authored by the first author investigated COVID--19 misinformation detection \cite{sikosana2024hybrid}, while Safarnejad et al.\ \cite{safarnejad2020contrasting} examined the dissemination networks of misinformation and accurate information during the 2016 Zika outbreak. Nevertheless, a rapid review by Chowdhury et al.\ \cite{chowdhury2023understanding} identified recurring misinformation themes across public health emergencies, including vaccine-related concerns and declining trust in health authorities. These recurring themes demonstrate the need for comparative research across outbreaks and for public health responses tailored to the characteristics of particular misinformation narratives and affected populations.

To address this gap, our study employs \textbf{BERTopic-VP} to conduct the first comparative, engagement-aware analysis of pandemic misinformation across COVID--19 and mpox twitter discourse. Unlike earlier approaches that treat themes in isolation, BERTopic-VP enables dynamic, cross-contextual tracking of misinformation topics while prioritising those with high virality potential. This supports scalable detection of both persistent and emergent narratives and provides actionable insight into how misinformation mutates in different sociopolitical contexts. For example, while COVID--19 misinformation focused on conspiracy theories and pseudoscience, mpox discourse centred on policy failures and stigma—insights surfaced through both topic modelling and virality-aware ranking.

These contrasting cases reinforce that misinformation is not merely a product of content but of context, shaped by evolving public trust, institutional response, and audience fatigue. BERTopic-VP allows us to quantify these shifts and prioritise responses accordingly, supporting evidence-based strategies in digital health communication. The findings illustrate the need for agile, event-sensitive misinformation surveillance tools that go beyond static modelling and respond to virality as it unfolds.

\subsection{Scalability \& processing efficiency of BERTopic-VP}
\label{sec:scalability}

To assess computational feasibility, we benchmarked the BERTopic--VP pipeline on fixed-size batches of 10{,}000 to 100{,}000 tweets. On a desktop workstation (Intel i7 CPU, 16GB RAM, NVIDIA RTX 3060 GPU), the end-to-end run (semantic embedding, UMAP dimensionality reduction, HDBSCAN clustering, and virality-aware enrichment) processed 10{,}000 tweets in under 60 seconds under the stated software and batching configuration. In this paper, ``near real-time'' is used in the limited sense of \emph{minute-to-hour} batch refresh, not sub-second streaming latency. This benchmark is reported as an indicative throughput figure for offline or periodic monitoring on commodity hardware, rather than as a guarantee of live, platform-integrated performance, since end-to-end latency in practice also depends on ingestion, storage, batching, and refresh cadence.

Prior work suggests that classical topic models such as LDA and NMF can produce less interpretable topics on short-text corpora, whereas embedding-based approaches often yield more coherent and granular groupings under similar sparsity constraints \cite{egger2022topic,grootendorst2022bertopic}. However, topic quality and coherence are not directly comparable across studies because they depend on the corpus, preprocessing, topic representation, and the specific coherence metric used. We therefore treat external LDA and NMF findings as contextual motivation rather than as a head-to-head benchmark against our experiments.

BERTopic-VP extends standard BERTopic by incorporating virality-aware prioritisation into the clustering and monitoring workflow. Using SBERT embeddings, UMAP dimensionality reduction, and HDBSCAN clustering, the pipeline adapts to shifting narratives without requiring the number of topics to be predefined. This is valuable in dynamic, high-volume social media settings. As noted by Grootendorst \cite{grootendorst2022bertopic}, updating the c-TF-IDF representations for new time windows enables longitudinal topic tracking without destabilising cluster structure.

In our implementation, we applied the same BERTopic-VP configuration to both COVID--19\_FNIR and Monkeypox corpora despite differences in time period and thematic content. The pipeline produced semantically interpretable topics without dataset-specific parameter changes, consistent with the embedding and density-based design of BERTopic \cite{grootendorst2022bertopic}. Coherence values on short microtext were modest, with mean $C_v$ of 0.084 for COVID--19\_FNIR dataset and 0.286 for Monkeypox dataset (Table~\ref{tab:coherence_bertopic}). This is consistent with coherence compression in sparse short-text settings. Higher coherence values reported elsewhere (for example 0.52 and 0.53) likely reflect more thematically uniform corpora and different preprocessing and aggregation choices \cite{egger2022topic}.

The VP module enables prioritisation of emergent high-impact content using a unified tweet-level prioritisation signal $\tilde{E}_i$. When engagement metadata are observed, $\tilde{E}_i$ is computed from realised interaction counts (for example likes and retweets), and tweets in the top $x \in \{1,5,10\}\%$ of the dataset-specific $\tilde{E}$ distribution are labelled \textit{viral}. When engagement metadata are unavailable, $\tilde{E}_i$ is given by the transferred propensity score (logit $\eta_i$), so the same percentile thresholds quantify estimated diffusion propensity rather than realised engagement. Cluster prioritisation is based on the cluster-level aggregate $v_j$ (Equation~\ref{eq:cluster_virality}), while the viral density $V(C_j)$ (Equation~\ref{eq:viral_density}) is reported as an auxiliary diagnostic of whether high-impact activity is broadly distributed across a topic or concentrated in a small subset of posts. This elevation mechanism highlights clusters that may be low volume but exhibit early signals of diffusion potential or persuasive traction, supporting timely analyst review for counter-messaging or fact-checking.

Overall, BERTopic-VP offers a scalable, virality-aware framework for health misinformation surveillance, capable of handling evolving discourse with modest computational overhead and analyst-oriented interpretability. Although the present evaluation uses static datasets, the architecture supports longitudinal tracking. Future work will visualise topic evolution across outbreak phases and quantify lead--lag relationships between virality signals and public health announcements to assess predictive potential.

\section{Materials and Methods}
\label{sec:materials_methods}
We developed a modular and scalable informatics pipeline, \textbf{BERTopic-VP}, for the detection, interpretation, and prioritisation of health misinformation narratives on Twitter. BERTopic-VP extends BERTopic by adding a virality prioritisation (VP) overlay that ranks discovered topic clusters using engagement-derived signals where available, and a transferred proxy signal otherwise. As shown in Figure~\ref{fig:end_to_end_vp_pipeline}, the pipeline is monitoring-oriented and supports periodic refresh cycles for dynamic threat triage, while this paper evaluates the framework offline on three static datasets (i.e., COVID--19\_FNIR \cite{saenz2021covid}, Constraint \cite{patwa2021fighting}, and Monkeypox  \cite{crone2022monkeypox}). The topic discovery module is applied using a fixed configuration across datasets, without dataset-specific retuning, while the VP signal uses observed engagement where available and a transferred proxy prioritisation score otherwise.

\subsubsection{Data split and evaluation protocol}
\label{sec:split_protocol}

For each dataset, we used a fixed, stratified 80/20 train--test split (stratified by the misinformation label) with a fixed random seed.
All model selection steps used five-fold cross-validation on the training split only, and final metrics are reported on the held-out 20\% test split.

\paragraph{Randomness control and stability.}
To support reproducibility, we fix random seeds for all stochastic components (data split, embedding projection, and clustering initialisation where applicable). Given the stochastic nature of UMAP and HDBSCAN, we additionally assess clustering stability by rerunning the pipeline multiple times under identical configurations and computing Adjusted Rand Index (ARI) between runs. We further evaluate the consistency of the top-$k$ VP-ranked clusters across runs, treating deviations as an indicator of sensitivity to stochastic initialisation.

\noindent\textbf{Verification resources.}
For the verification channel, outbreak-specific repositories
\(\mathcal{R}_{\text{COVID}}\) and \(\mathcal{R}_{\text{MPX}}\) are
used, as described in Section~\ref{sec:outbreak_verification_resources}

\paragraph{Engagement availability across datasets.}
Only the Monkeypox dataset contains \emph{observed} tweet-level engagement fields (e.g., likes, retweets, and replies).
COVID--19\_FNIR and Constraint do not provide native engagement metadata in their released schemas, so VP cannot use realised interaction counts for those datasets.
Accordingly, for COVID--19\_FNIR and Constraint, VP operates on a \emph{proxy} prioritisation signal derived from a propensity-to-spread model fitted on Monkeypox dataset and transferred as a ranking score.
Throughout this paper, any VP outputs for COVID--19\_FNIR and Constraint should therefore be interpreted as \emph{estimated diffusion propensity}, not observed platform engagement.
Formal definitions of $E_i$ and the unified prioritisation signal $\tilde{E}_i$ are given in Equations~\ref{eq:engagement_observed} and \ref{eq:engagement_surrogate}.

Table~\ref{tab:engagement_availability} summarises the engagement fields available in each corpus and the VP signal used throughout this paper. Formal definitions of $E_i$ and the unified prioritisation signal $\tilde{E}_i$ are given in Equations~\ref{eq:engagement_observed} and \ref{eq:engagement_surrogate}.

\begin{table*}[!t]
\centering

\caption{Engagement metadata availability and VP signal used in each dataset.}
\label{tab:engagement_availability}

\small
\setlength{\tabcolsep}{5pt}
\renewcommand{\arraystretch}{1.15}

\begin{tabularx}{\textwidth}{
@{}
>{\raggedright\arraybackslash}p{2.6cm}
>{\centering\arraybackslash}p{2.8cm}
>{\centering\arraybackslash}p{4.2cm}
>{\raggedright\arraybackslash}X
@{}
}
\toprule

\textbf{Dataset} &
\textbf{Observed engagement?} &
\textbf{VP signal used} &
\textbf{Interpretation} \\

\midrule

Monkeypox &
Yes &
Observed \(E_i\) &
Realised interaction counts drive VP ranking. \\

COVID--19\_FNIR &
No &
Proxy \(\tilde{E}_i\) from \(\eta_i\), equivalently \(\hat{p}_i\) &
VP ranking reflects predicted diffusion propensity, not realised engagement. \\

Constraint &
No &
Proxy \(\tilde{E}_i\) from \(\eta_i\), equivalently \(\hat{p}_i\) &
VP ranking reflects predicted diffusion propensity, not realised engagement. \\

\bottomrule
\end{tabularx}

\vspace{2pt}

\begin{minipage}{0.97\textwidth}
\footnotesize
\raggedright
\noindent
\textit{Note.}
Monkeypox uses observed engagement \(E_i\) computed from the available interaction fields. COVID--19\_FNIR and Constraint use a proxy prioritisation score \(\tilde{E}_i\), derived from the propensity-model score \(\eta_i\), equivalently \(\hat{p}_i\), because native engagement metadata are unavailable.
\end{minipage}

\end{table*}

\subsection{Topic Modelling via BERTopic-VP}

We applied the BERTopic-VP pipeline separately to each dataset. At its core, BERTopic \cite{grootendorst2022bertopic} combines sentence-transformer embeddings with Uniform Manifold Approximation and Projection (UMAP) \cite{mcinnes2018umap} for dimensionality reduction and Hierarchical Density-Based Spatial Clustering of Applications with Noise (HDBSCAN) \cite{campello2013density} for clustering. This architecture is well-suited to short, noisy texts such as tweets, where embedding-based topic models often yield more coherent and fine-grained themes than classical bag-of-words approaches in comparable microtext settings, and where LDA \cite{Blei2003LDA} and NMF \cite{Lee1999NMF} can be sensitive to sparsity and vocabulary fragmentation.

We used the \texttt{all-MiniLM-L6-v2} Sentence-BERT encoder (uncased, 384-dimensional embeddings) to generate contextual sentence representations.

To transform BERTopic into BERTopic-VP, we extend the pipeline by computing post hoc engagement scores for each tweet (likes $l_i$, retweets $r_i$, and replies $q_i$ where available). Where native engagement is unavailable, we use the propensity-to-spread proxy defined in Step~\ref{step:propensity_proxy}. These signals are then aggregated at the cluster level, allowing the system to prioritise clusters exhibiting high diffusion propensity.

\paragraph{Topic labelling protocol.}
Topic labels are assigned as descriptive short-hands for reporting. For each cluster $C_j$, we inspect the top terms from the topic representation and a fixed number of exemplar posts sampled from high-probability members. Labels are only used when the cluster shows a clear and stable semantic core; otherwise, the cluster is reported using its top terms without a narrative title. This reduces the risk of over-interpreting ambiguous micro-topics.

\noindent\textbf{Reproducibility.}
Unless stated otherwise, we report all primary results using a fixed preprocessing pipeline and fixed random seeds for stochastic components (UMAP and HDBSCAN) to support reproducibility. For Constraint only, we additionally report a rerun-sensitivity check by varying the UMAP random seed over $R=10$ reruns while keeping preprocessing, embeddings, and HDBSCAN hyperparameters fixed (Section~\ref{sec:constraint_instability}, Table~\ref{tab:constraint_stability}).

\noindent\textbf{Model hyperparameters.}
UMAP and HDBSCAN were configured using fixed hyperparameters for the final BERTopic-VP pipeline, with UMAP controlling the local structure and dimensionality of the embedding projection, and HDBSCAN controlling cluster density sensitivity and minimum cluster size. The final settings are summarised in Table~\ref{tab:umap_hdbscan_config}.

\begin{table}[!t]
\centering
\caption{UMAP and HDBSCAN configuration used for BERTopic-VP.}
\label{tab:umap_hdbscan_config}
\footnotesize
\setlength{\tabcolsep}{5pt}
\renewcommand{\arraystretch}{1.15}
\begin{tabular}{lll}
\toprule
\textbf{Component} & \textbf{Hyperparameter} & \textbf{Value} \\
\midrule
UMAP & \texttt{n\_neighbors} & 15 \\
UMAP & \texttt{min\_dist} & 0.1 \\
UMAP & \texttt{n\_components} & 5 \\
HDBSCAN & \texttt{min\_cluster\_size} & 50 \\
HDBSCAN & \texttt{min\_samples} & 10 \\
\bottomrule
\end{tabular}
\end{table}

\subsubsection{Hyperparameter selection}
UMAP was held fixed as shown in Table~\ref{tab:umap_hdbscan_config}. HDBSCAN hyperparameters were selected \emph{once} on a held-out development split via a small grid search, \texttt{min\_cluster\_size} $\in \{20,50,80\}$ and \texttt{min\_samples} $\in \{5,10,20\}$, using topic coherence ($C_v$) as the primary criterion alongside a stable cluster count and a bounded noise rate. The selected configuration, \texttt{min\_cluster\_size}=50 and \texttt{min\_samples}=10 (Table~\ref{tab:umap_hdbscan_config}), was then held fixed and applied unchanged across COVID--19\_FNIR, Monkeypox, and Constraint datasets, without dataset-specific retuning.

\subsection{Pipeline Steps}
\label{sec:pipeline_steps}
Figure~\ref{fig:end_to_end_vp_pipeline} presents the \textbf{BERTopic-VP} pipeline, integrating semantic topic modelling, engagement-informed virality scoring, misinformation detection, thematic evaluation, and scalability considerations.

\begin{enumerate}

\item \textbf{Data Ingestion and Preprocessing.}\par
\begin{equation}
T = \{ t_1, t_2, \ldots, t_n \}
\label{eq:data_set}
\end{equation}
Each tweet $t_i$ is associated with metadata $M(t_i)$. Preprocessing maps raw tweets into cleaned and embedded form:
\begin{equation}
P: T \times M \rightarrow T'
\label{eq:preprocessing}
\end{equation}

\item \textbf{Semantic Embedding, Dimensionality Reduction, and Clustering.}\par
Tweets are embedded, reduced, and clustered as follows:
\begin{align}
E(T') &= f_{\text{BERT}}(T') \label{eq:embedding} \\
U(T') &= f_{\text{UMAP}}(E(T')) \label{eq:umap} \\
C(T') &= f_{\text{HDBSCAN}}(U(T')) \label{eq:hdbscan}
\end{align}
The induced topic partition is:
\begin{equation}
\mathcal{C} = \{ C_1, \ldots, C_k \}, \quad \bigcup_{j=1}^k C_j = T'.
\label{eq:topic_set}
\end{equation}

\item \textbf{Engagement Metric Extraction \& Surrogate Propensity.} \label{step:propensity_proxy} \par
For datasets with observed engagement (e.g., Monkeypox dataset), the engagement of tweet $t_i$ is represented by:
\begin{equation}
E_i = l_i + r_i + q_i + s_i \; (+ \; c_i \text{ where available}),
\label{eq:engagement_observed}
\end{equation}
where $l_i$, $r_i$, $q_i$, $s_i$, and $c_i$ denote likes, retweets, replies, quotes, and link clicks, respectively. When a component is absent for a given corpus (that is, the field is not recorded in the dataset schema), the composite is computed over the observed components only, and the missing component is treated as \emph{unobserved} rather than behaviourally equal to zero.

In this paper we use equal aggregation weights (equivalently $\mathbf{w}=\mathbf{1}$), so $E_i$ reduces to the unweighted sum in Equation~\ref{eq:engagement_observed}. The adaptive update in Section~\ref{sec:feedback_loop} describes how weights and thresholds can be recalibrated in a streaming deployment.

For datasets without engagement metadata (COVID--19\_FNIR and Constraint), we compute a propensity-to-spread score $\hat{p}_i \in (0,1)$ using a logistic regression model trained on Monkeypox dataset. The model predicts whether a tweet belongs to a high-engagement regime (top-$q$ quantile of $E_i$) from text-derived and psychologically motivated features available across datasets. In these corpora, $\hat{p}_i$ is used as a surrogate virality signal.

We set $q=0.10$ (top 10\%) to define the high-engagement regime during propensity model training on the Monkeypox dataset, using observed engagement $E_i$. This choice yields a sparse but learnable positive class, and aligns with the long-tailed nature of engagement distributions on social media.

We denote the model’s linear score by $\eta_i=\mathbf{w}^{\top}\mathbf{z}_i+b$, with the corresponding probability $\hat{p}_i=\sigma(\eta_i)$. In VP ranking and in Table~\ref{tab:vp_results}, we report $\eta_i$ because it preserves score separation in the high-propensity regime.

\paragraph{Proxy validation on Monkeypox dataset (within-domain check).}
Because the propensity model is trained on Monkeypox dataset using observed $E_i$, we first perform a within-domain validation check before transferring the score to datasets without engagement data. On the Monkeypox dataset test split, we evaluate (i) ROC--AUC for predicting the top-$q\%$ high-engagement regime, and (ii) the rank correlation between $\eta_i$ and observed $E_i$ using Spearman’s $\rho$. Although this does not guarantee cross-event calibration, it shows that $\eta_i$ provides a meaningful signal for ordering engagement within the domain in which it is learned. This supports its use as an ordinal prioritisation signal when only ranking is required.

\textbf{Transfer assumption and interpretation.}
When applied to datasets without engagement data, $\hat{p}_i$ is interpreted as an ordinal estimate of propensity to spread, learned from Monkeypox dataset (Step~\ref{step:propensity_proxy}), rather than as a direct measure of engagement or virality. Therefore, cross-dataset comparisons in this paper are based on within-dataset ranks and percentile thresholds of $\tilde{E}$, rather than assuming that values of $E_i$ and $\hat{p}_i$ are directly comparable in magnitude.

\noindent\textbf{Rationale for transfer.}
The transfer assumption is not that absolute engagement levels are comparable across outbreaks, but that the \emph{relative} cues associated with higher diffusion propensity are partially stable across health-misinformation settings.
In practice, the propensity model is used only to induce an \emph{ordering} of posts within each engagement-missing dataset, and VP operates exclusively through within-dataset percentiles of $\tilde{E}_i$ and $\{v_j\}_{j=1}^{k}$.
We therefore interpret $\eta_i$ (or $\hat{p}_i$) as an ordinal prioritisation signal rather than a calibrated estimate of realised interaction counts, and we avoid cross-dataset claims about engagement magnitude.

To unify notation across datasets, we define the prioritisation signal:
\begin{equation}
\tilde{E}_i=
\begin{cases}
E_i, & \text{if engagement metadata are available},\\
\eta_i, & \text{if engagement metadata are unavailable},
\end{cases}
\label{eq:engagement_surrogate}
\end{equation}
where $\eta_i=\mathbf{w}^{\top}\mathbf{z}_i+b$ is the propensity model logit and $\hat{p}_i=\sigma(\eta_i)$.
Because $\sigma(\cdot)$ is monotonic, ranking by $\eta_i$ is equivalent to ranking by $\hat{p}_i$.
We use $\eta_i$ to retain resolution among high-propensity posts, where $\hat{p}_i$ can saturate near 1.
Since $E_i$ and $\eta_i$ are used only through within-dataset ranks and percentile thresholds, no cross-dataset scale equivalence is assumed.

Since $E_i$ and $\eta_i$ are used only through within-dataset ranks and percentile thresholds, no cross-dataset scale equivalence is assumed.

\begin{equation}
v_j = \frac{1}{|C_j|} \sum_{t_i \in C_j} \tilde{E}_i
\label{eq:cluster_virality}
\end{equation}

For clarity, VP denotes \emph{Virality Prioritisation} throughout this paper. We use the term VP throughout to denote percentile-based prioritisation, and report $V(C_j)$ as the within-cluster share of VP-flagged tweets. For cluster-level screening, we compute a viral density score, defined as the within-cluster share of VP-flagged tweets:
\begin{equation}
V(C_j) = \frac{\left|\left\{t_i \in C_j \;\middle|\; t_i \text{ is VP-flagged}\right\}\right|}{|C_j|}.
\label{eq:viral_density}
\end{equation}
We report $V(C_j)$ as an auxiliary diagnostic of whether high-impact activity is broadly distributed across the cluster or concentrated in a small subset of posts. For downstream prioritisation, a tweet is considered \emph{VP-flagged} if $\tilde{E}_i$ lies within the top $x \in \{1,5,10\}$ percentiles of the dataset-specific $\tilde{E}_i$ distribution. A cluster is considered high priority if $v_j$ lies within the top $x$ percentiles of the $\{v_j\}_{j=1}^{k}$ distribution.

\item \textbf{Misinformation Detection \& Verification Fusion.}\par

Final misinformation probabilities are produced by fusing a content-based classifier with a verification channel:
\begin{equation}
p_{\mathrm{final}}(t_i) = \alpha\, p_{\mathrm{content}}(t_i) + (1-\alpha)\, p_{\mathrm{verify}}(t_i),
\label{eq:fusion}
\end{equation}
where $\alpha \in (0,1)$ controls the relative contribution of each channel. The parameter $\alpha$ is selected by grid search over $\{0.1,0.2,\ldots,0.9\}$ using validation folds on the training split, optimising macro-$F_1$ (and secondarily ROC--AUC). The best setting in our experiments was $\alpha=0.6$. Accordingly, the final model assigns 60\% weight to the content-based classifier and 40\% weight to the verification channel.

For downstream prioritisation, a tweet is considered \emph{VP-flagged} if $\tilde{E}_i$ lies within the top $x \in \{1,5,10\}$ percentiles of the dataset-specific $\tilde{E}_i$ distribution. A cluster is considered \emph{high priority} if $v_j$ lies within the top $x$ percentiles of the $\{v_j\}_{j=1}^{k}$ distribution.

\item \textbf{Thematic Cluster Evaluation.}\par
Topic clusters are evaluated using coherence ($C_v$), linguistic complexity indices, and fine-grained emotion profiles.
To connect thematic outputs to VP, we use percentile thresholding under the prioritisation signal $\tilde{E}_i$ and report the viral density $V(C_j)$ in Eq.~\ref{eq:viral_density}, defined as the within-cluster share of VP-flagged tweets (top-$x\%$ under $\tilde{E}_i$ for $x\in\{1,5,10\}$).

\item \textbf{Adaptive feedback loop for VP recalibration (deployment logic).}\par
This component is described for completeness as part of the end-to-end monitoring design. The experiments in this paper report offline results under a fixed configuration and a single-pass scoring of $\tilde{E}_i$ (see Section~\ref{sec:feedback_loop}).

\end{enumerate}

\subsection{Outbreak-specific verification resources}
\label{sec:outbreak_verification_resources}

To avoid domain mismatch across outbreaks, the verification channel conditions on an outbreak-specific reference set.
Let $d(t_i)\in\{\text{COVID},\text{MPX}\}$ denote the outbreak domain of tweet $t_i$, and let $\mathcal{R}_{d(t_i)}$ denote the corresponding public-health verification repository.

For COVID--19 corpora, $\mathcal{R}_{\text{COVID}}$ is instantiated using the WHO COVID--19 Mythbusters resource \cite{who_mythbusters_covid}.
For mpox corpora, $\mathcal{R}_{\text{MPX}}$ is instantiated using outbreak-specific public-health resources, including WHO mpox myth-busting and guidance materials, and complementary guidance pages from agencies such as the CDC and UKHSA  \cite{who_mpox,cdc_mpox,ukhsa_mpox}.
Claim-veracity signals are obtained by matching tweet claims to the most relevant entries in $\mathcal{R}_{d(t_i)}$ (using the same matching procedure across domains), then combining these signals with source-credibility indicators to form $p_{\text{verify}}(t_i)$.

\noindent\textbf{Claim extraction and repository matching.}
For verification, the tweet text is treated as the claim span after
standard cleaning, including lowercasing, URL and mention removal, and
whitespace normalisation. Let \(q_i\) denote the cleaned claim text for
tweet \(t_i\), and let
\(\mathcal{R}_{d(t_i)}=\{r_1,\ldots,r_m\}\) denote the outbreak-specific
repository entries defined in this subsection.

Two-stage matching is performed: (i) lexical candidate retrieval using
BM25 over \(\mathcal{R}_{d(t_i)}\) to obtain the top-\(K\) candidates;
and (ii) semantic re-ranking using cosine similarity between
Sentence-BERT embeddings of \(q_i\) and each candidate entry. We use
\(K=10\) and \(\tau=0.60\) in all experiments. The highest-scoring entry
is selected, and a match is accepted only when semantic similarity
exceeds \(\tau\); otherwise, the claim is treated as having no reliable
match and the verification signal relies on source-credibility features
alone.

\noindent\textbf{Calibration.}
The raw verification output is calibrated to a probability using isotonic regression on the training split via five-fold cross-validation.
For each fold, isotonic regression is fit on the fold’s validation predictions and applied to obtain calibrated $p_{\text{verify}}(t_i)$.
The calibrated verification probability is then fused with the content classifier via Eq.~\ref{eq:fusion}, and final metrics are reported on the held-out test split.

Raw verification scores are calibrated on held-out validation folds using isotonic regression to improve probabilistic reliability, ensuring that $p_{\text{verify}}$ behaves as a calibrated probability estimate prior to fusion.

\subsubsection{Adaptive feedback loop for VP recalibration}
\label{sec:feedback_loop}
To maintain stable prioritisation under shifting discourse, the end-to-end virality framework applies an adaptive feedback loop at each refresh cycle. Let $\Delta t$ denote the refresh interval and let $\mathcal{W}_t$ denote the sliding window used at cycle $t$.\par

\textbf{Fixed components.} Topic discovery (embedding model, UMAP, and HDBSCAN hyperparameters) is held fixed once selected and applied consistently across datasets without dataset-specific retuning.\par

\textbf{Updated components.} At each cycle $t$, the loop updates (i) virality aggregation weights used to combine engagement channels into $E_i$ when native engagement is observed, and (ii) the prioritisation threshold $\tau_t$ used to flag high-risk posts or clusters. In this paper, equal aggregation weights are used (equivalently $\mathbf{w}=\mathbf{1}$), so $E_i$ reduces to the unweighted composite in Equation~\ref{eq:engagement_observed}. For datasets without native engagement metadata, thresholding operates on the proxy signal $\tilde{E}_i$ derived from the propensity model logit $\eta_i$ (equivalently $\hat{p}_i$) rather than on realised interaction counts.\par

\textbf{Update rule.} Given engagement channels $g_{i,1},\dots,g_{i,K}$ for post $i$, an instantaneous weight estimate $\hat{\mathbf{w}}_t$ is computed on $\mathcal{W}_t$ and smoothed via
\[
\mathbf{w}_t = (1-\lambda)\mathbf{w}_{t-1} + \lambda \hat{\mathbf{w}}_t,
\]
where $\lambda \in (0,1]$ controls responsiveness. The prioritisation threshold is set by a quantile rule on the current score distribution:
\[
\tau_t = Q_{q_{\mathrm{VP}}}\big(\tilde{E}_i \mid i \in \mathcal{W}_t\big).
\]
\noindent In offline evaluation, VP thresholding uses $x\in\{1,5,10\}\%$ (so $q_{\mathrm{VP}}=x/100$), whereas propensity training uses $q=0.10$ to define the high-engagement class on Monkeypox dataset.

In this paper, we report offline results and do not execute online recalibration cycles, so $\tau_t$ and $\mathbf{w}_t$ are presented as deployment logic rather than applied updates. Under streaming deployment, this rule would flag approximately the top $q$ fraction of posts under the current regime.\par

\paragraph{Optional normalised and rate-based extensions (not used in this paper).}
Where richer metadata are available, VP can be extended beyond raw engagement counts to reduce comparability bias across accounts. For example, an audience-normalised engagement score can be defined as
\begin{equation}
E_i^{(\mathrm{norm})} = \frac{E_i}{1+\log\!\big(1+F_i\big)},
\label{eq:engagement_normalised}
\end{equation}
where $F_i$ denotes the author follower count. A simple time-normalised rate can be defined using the tweet timestamp as
\begin{equation}
E_i^{(\mathrm{rate})} = \frac{E_i}{1+\Delta t_i},
\label{eq:engagement_rate}
\end{equation}
where $\Delta t_i$ is the elapsed time (in hours) between posting and observation. \textit{These extensions are not applied in this paper} because follower and timing fields are not available consistently across COVID--19\_FNIR, Constraint, and Monkeypox dataset, and engagement timestamps required for true diffusion velocity are not provided in the released schemas.

\subsection{\textbf{Scalability \& monitoring-oriented design.}}\par
The pipeline supports batch-based processing with configurable refresh cadence (for example, hourly or daily re-runs), which is suitable for surveillance-style monitoring where periodic updates are acceptable. Its modular structure allows the embedding, reduction, clustering, and enrichment stages to be profiled and scaled independently. Transformer-based encoders produce contextual embeddings, UMAP reduces dimensionality to support efficient density-based clustering, and HDBSCAN forms clusters without requiring a pre-specified number of topics. In this paper, we evaluate the approach in an offline setting on static datasets, and we frame any streaming or incremental incorporation of new data as an architectural capability that would require an explicit online implementation and operational constraints.

\subsection{Methodological Summary}
BERTopic-VP is a scalable, engagement-aware pipeline for health misinformation surveillance. It couples semantically rich topic modelling with virality-informed prioritisation, supporting monitoring-oriented periodic refresh cycles for analyst triage and early narrative surfacing.

\begin{figure*}[htbp]
    \centering
    \includegraphics[width=0.85\textwidth]{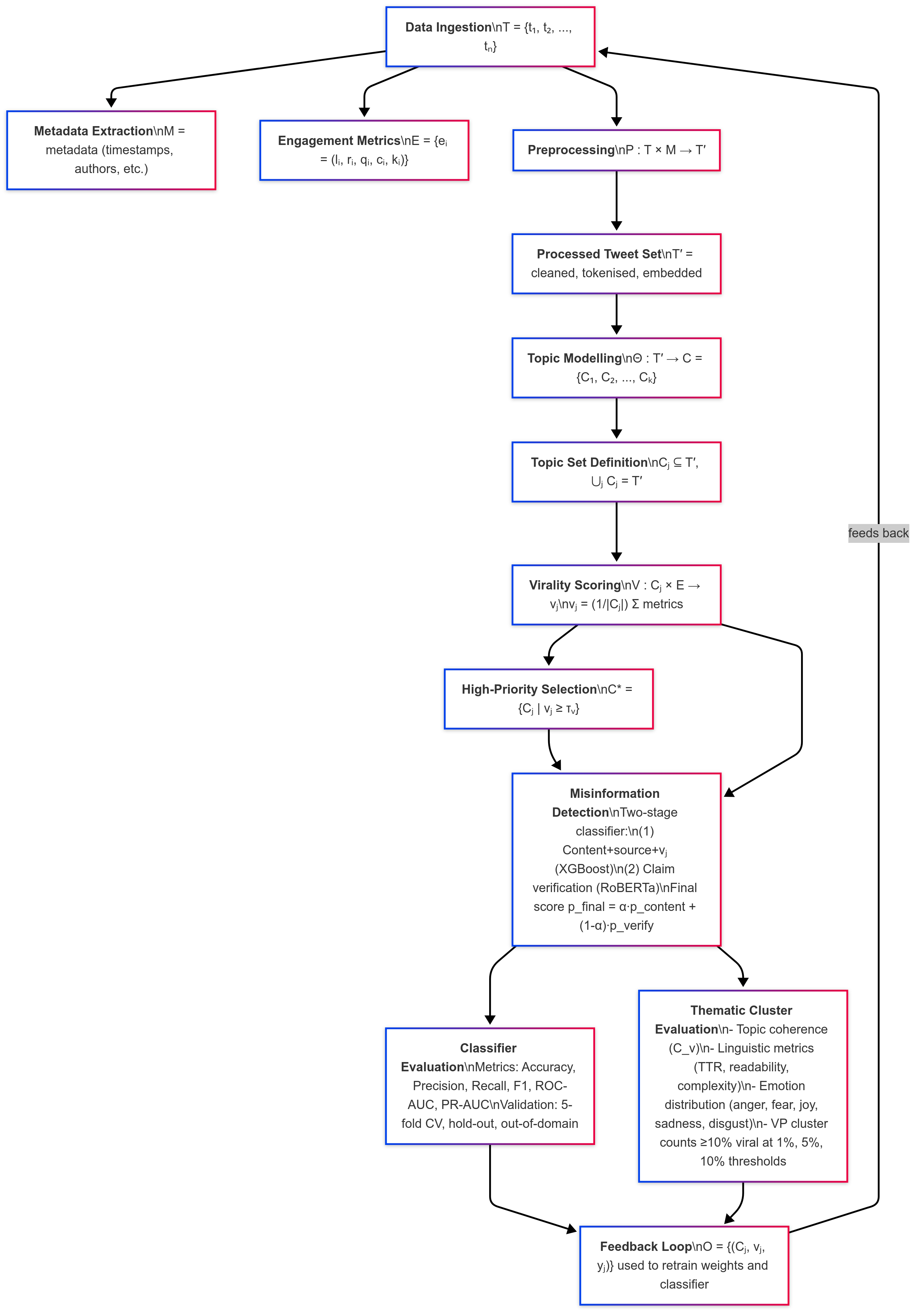}
    \caption{\textbf{End-to-End Virality-Prioritised Misinformation Detection Pipeline.}}
    \label{fig:end_to_end_vp_pipeline}
\end{figure*}

\section{Results}
\label{sec:results}

We applied the \textbf{BERTopic-VP} pipeline to three misinformation corpora, COVID--19\_FNIR, Constraint and Monkeypox datasets. We report unsupervised clustering quality, downstream supervised classification, and VP outcomes. Results are presented comparatively across datasets.

\noindent\textbf{Constraint stability note.}
Constraint dataset cluster partitions exhibit higher sensitivity to sampling variation and HDBSCAN noise assignment than COVID--19\_FNIR and Monkeypox datasets, so threshold-based VP prevalence is not reported quantitatively for Constraint; instead, we report qualitative VP case studies for interpretability, and treat Constraint cluster metrics as descriptive rather than directly comparable (see Section~\ref{sec:constraint_instability} and Table~\ref{tab:constraint_stability}).

\subsection{Unsupervised Clustering Performance}
\label{subsec:unsupervised}
Table~\ref{tab:clustering_results} summarises HDBSCAN quality metrics, including Normalized Mutual Information (NMI), Adjusted Rand Index (ARI), Purity, cluster count, and topic coherence ($C_v$). ARI evaluates pairwise agreement between clustering assignments and true labels, adjusted for chance. NMI measures the amount of shared information between the generated clusters and ground truth. Purity assesses the extent to which each cluster contains data from a single dominant class. Coherence ($C_v$) evaluates the semantic interpretability of the clusters \cite{lee2025semi,meaney2023quality}. For interpretive clarity, these metrics may also be understood as addressing the following questions: \textbf{ARI:} \textit{Are pairs of items grouped together or separated in a way that matches the real labels, beyond random chance?} \textbf{NMI:} \textit{How well does the overall clustering structure correspond to the true class structure?} \textbf{Purity:} \textit{When HDBSCAN groups items together, how often are those items mostly from the same known class?} \textbf{Coherence:} \textit{Do the words defining each cluster form meaningful and interpretable themes?}

\textbf{Monkeypox dataset} achieved the highest purity and NMI, together with a moderate average coherence. COVID--19\_FNIR showed moderate purity with the lowest coherence, which is consistent with a noisier and more heterogeneous narrative space. Constraint obtained the lowest purity, however its average coherence was the highest, indicating tight intra-topic semantics that align less with the available labels. 

\begin{table*}[!t]
\centering
\caption{Comparative unsupervised topic clustering performance under BERTopic-VP. The number of clusters corresponds to the final non-outlier topic count used as the VP denominator.}
\label{tab:clustering_results}

\small
\setlength{\tabcolsep}{8pt}
\renewcommand{\arraystretch}{1.12}

\begin{tabularx}{\textwidth}{
@{}
>{\raggedright\arraybackslash}X
>{\centering\arraybackslash}p{2.0cm}
>{\centering\arraybackslash}p{2.0cm}
>{\centering\arraybackslash}p{2.0cm}
>{\centering\arraybackslash}p{2.2cm}
>{\centering\arraybackslash}p{2.8cm}
@{}
}
\toprule

\textbf{Dataset} &
\textbf{Purity} &
\textbf{NMI} &
\textbf{ARI} &
\textbf{\# Clusters} &
\textbf{Coherence (\(C_v\))} \\

\midrule

COVID--19\_FNIR
& 0.742
& 0.210
& 0.181
& 112
& 0.084 \\

Monkeypox
& \textbf{0.940}
& \textbf{0.281}
& \textbf{0.233}
& 87
& 0.286 \\

Constraint
& 0.522
& 0.036
& -0.002
& 5
& \textbf{0.463} \\

\bottomrule
\end{tabularx}

\vspace{2pt}

\begin{minipage}{0.97\textwidth}
\footnotesize
\raggedright
\noindent
\textit{Note.}
Purity, NMI, and ARI quantify alignment between discovered topics and available labels, and are therefore sensitive to class definitions and topic granularity. The reported cluster count excludes outliers assigned to HDBSCAN noise and represents the denominator used in VP thresholding. The ARI reported here is a label-alignment metric, comparing discovered topics with class labels, and is distinct from the rerun-stability ARI reported for Constraint in Table~\ref{tab:constraint_stability}. For Constraint, the final non-outlier topic count reported here corresponds to the main fixed run used for the clustering metrics, whereas rerun sensitivity is assessed separately under a tractable rerun setting in Table~\ref{tab:constraint_stability}.
\end{minipage}

\end{table*}
For COVID--19\_FNIR, the fitted model yielded a relatively fine-grained partition (112 topics) but a modest mean coherence ($C_v = 0.084$), which is consistent with coherence compression in sparse short-text settings (Table~\ref{tab:coherence_bertopic}).

\subsection{Supervised Misinformation Detection}
\label{subsec:supervised}
We trained the two-stage classifier on the training split and selected hyperparameters using five-fold cross-validation within the training data.
We then report final performance on the held-out test split.
Performance was strong across domains, with ROC--AUC and PR--AUC above 0.95 in all cases.
Monkeypox dataset produced the highest accuracy and F1, followed by COVID--19\_FNIR, then Constraint.
Detailed metrics are given in Table~\ref{tab:classification_results}.

\begin{table*}[!t]
\centering
\caption{Comparative Supervised Classification Performance}
\label{tab:classification_results}

\small
\setlength{\tabcolsep}{8pt}
\renewcommand{\arraystretch}{1.12}

\begin{tabularx}{\textwidth}{
@{}
>{\raggedright\arraybackslash}X
>{\centering\arraybackslash}p{2.1cm}
>{\centering\arraybackslash}p{2.1cm}
>{\centering\arraybackslash}p{1.9cm}
>{\centering\arraybackslash}p{1.8cm}
>{\centering\arraybackslash}p{2.1cm}
>{\centering\arraybackslash}p{2.0cm}
@{}
}
\toprule

\textbf{Dataset} &
\textbf{Accuracy} &
\textbf{Precision} &
\textbf{Recall} &
\textbf{F1} &
\textbf{ROC-AUC} &
\textbf{PR-AUC} \\

\midrule

COVID--19\_FNIR
& 0.906
& 0.897
& 0.918
& 0.907
& 0.966
& 0.961 \\

\textbf{Monkeypox}
& \textbf{0.956}
& \textbf{0.937}
& \textbf{0.963}
& \textbf{0.950}
& \textbf{0.989}
& \textbf{0.985} \\

Constraint
& 0.895
& 0.901
& 0.897
& 0.899
& 0.951
& 0.953 \\

\bottomrule
\end{tabularx}

\vspace{2pt}

\begin{minipage}{0.97\textwidth}
\footnotesize
\raggedright
\noindent
\textit{Note.}
Metrics are reported on the held-out test split under a fixed preprocessing and training protocol, with model selection performed via five-fold cross-validation on the training split. PR--AUC denotes the area under the precision--recall curve and is reported alongside ROC--AUC because it is informative under class imbalance.
\end{minipage}

\end{table*}

\subsection{Emotional Tone, Lexical Simplicity, \& Engagement Alignment}
\label{sec:emotional_lexical_engagement}
As summarised in Table~\ref{tab:sentiment_emotion} and Table~\ref{tab:linguistic_metrics}, analysis of COVID--19\_FNIR, Mpox, and Constraint tweets indicates that emotionally charged language and simpler lexical style are consistent correlates of VP prioritisation within the \textbf{BERTopic-VP} framework. Here, “prioritisation” refers to the unified VP signal $\tilde{E}_i$, which equals observed engagement $E_i$ where available (Monkeypox dataset) and otherwise a proxy propensity score (COVID--19\_FNIR and Constraint). Across datasets, clusters combining higher emotional intensity with more accessible language were more likely to be prioritised by VP, even when cluster volume was modest, consistent with evidence that affectively arousing and cognitively light content diffuses readily in online settings.

\begin{table*}[!t]
\centering
\caption{Average sentiment polarity and emotion scores by pandemic source.}
\label{tab:sentiment_emotion}

\small
\setlength{\tabcolsep}{8pt}
\renewcommand{\arraystretch}{1.15}

\begin{tabularx}{\textwidth}{
@{}
>{\raggedright\arraybackslash}X
>{\centering\arraybackslash}p{1.8cm}
>{\centering\arraybackslash}p{1.5cm}
>{\centering\arraybackslash}p{1.5cm}
>{\centering\arraybackslash}p{1.5cm}
>{\centering\arraybackslash}p{1.7cm}
@{}
}
\toprule

\textbf{Pandemic Source} &
\textbf{Polarity} &
\textbf{Joy} &
\textbf{Fear} &
\textbf{Anger} &
\textbf{Sadness} \\

\midrule

COVID--19\_FNIR
& 0.051
& 0.202
& 0.150
& 0.132
& 0.118 \\

Constraint
& 0.037
& 0.181
& 0.167
& 0.144
& 0.103 \\

Monkeypox
& 0.022
& 0.135
& 0.226
& 0.193
& 0.152 \\

\bottomrule
\end{tabularx}

\vspace{2pt}

\begin{minipage}{0.97\textwidth}
\footnotesize
\raggedright
\noindent
\textit{Note.}
Values are dataset-level means computed by averaging per-tweet polarity and emotion scores. Emotion scores are reported on the scale produced by the chosen extraction method and are intended for relative comparison across datasets under a consistent pipeline.
\end{minipage}

\end{table*}

\noindent\textit{Operationalisation.} Sentiment polarity was computed using \textsc{VADER}, while discrete emotion scores (Joy, Fear, Anger, Sadness) were extracted using the \textsc{NRC} emotion lexicon and averaged over tweets within each dataset. Lexical simplicity metrics reported in Table~\ref{tab:linguistic_metrics} were computed using \textsc{textstat} readability and lexical measures.

\begin{figure}[H]
    \centering
    \includegraphics[width=0.7\linewidth]{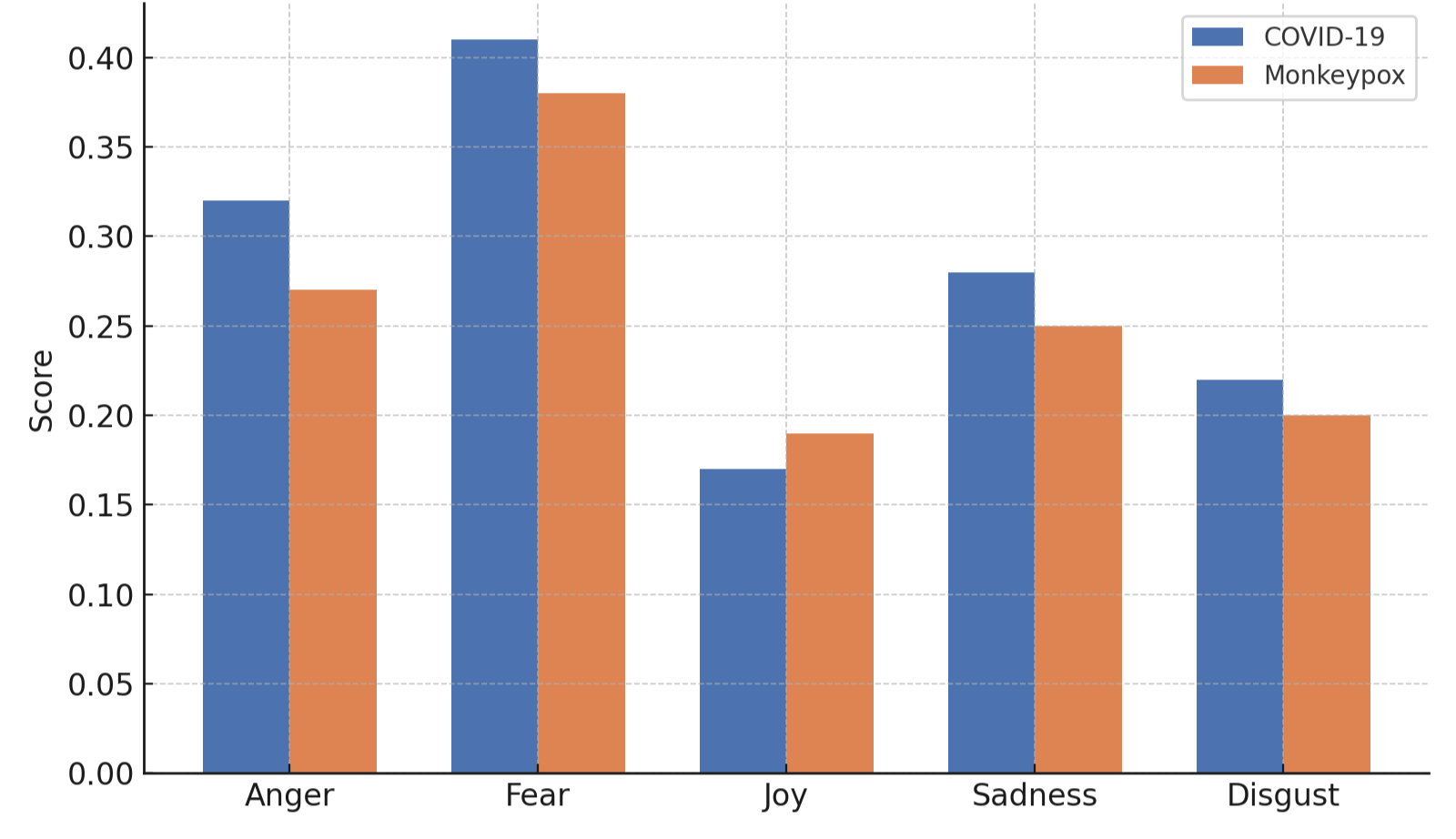}
    \caption{Distribution of emotion scores (anger, fear, joy, sadness, and disgust) across COVID--19\_FNIR and mpox misinformation tweets, highlighting divergent affective triggers.}
    \label{fig:emotion_scores}
\end{figure}

\textbf{Emotional tone} was quantified through sentiment polarity and discrete emotion scores. Mpox misinformation exhibited the most intense emotional activation, with notably higher fear (0.226), anger (0.193), and sadness (0.152) than COVID--19\_FNIR (fear 0.150, anger 0.132, sadness 0.118) and Constraint (fear 0.167, anger 0.144, sadness 0.103). These heightened affective signatures were most prominent in clusters that VP prioritised (high $\tilde{E}_i$ and high $v_j$),  particularly narratives criticising government response or amplifying outbreak severity. Figure~\ref{fig:emotion_scores} visualises these differences, showing higher fear- and anger-weighted emotion profiles for mpox relative to COVID--19\_FNIR.

\textbf{Lexical simplicity} was assessed using readability metrics (Table~\ref{tab:linguistic_metrics}). Mpox tweets scored higher on the Flesch Reading Ease (55.73 vs.\ COVID--19\_FNIR’s 30.26) and required a lower Flesch--Kincaid Grade Level (8.90 vs.\ 12.95), indicating greater accessibility. This lower cognitive load likely facilitated rapid message processing and sharing. In contrast, COVID--19\_FNIR misinformation was linguistically denser, with longer sentences and lower readability, possibly to simulate expertise but at the cost of immediate diffusion.

\begin{table*}[!t]
\centering
\caption{Average linguistic metrics for COVID--19\_FNIR and mpox misinformation tweets.}
\label{tab:linguistic_metrics}

\small
\setlength{\tabcolsep}{7pt}
\renewcommand{\arraystretch}{1.2}

\begin{tabularx}{\textwidth}{
@{}
>{\raggedright\arraybackslash}p{3.0cm}
>{\centering\arraybackslash}p{1.6cm}
>{\centering\arraybackslash}p{3.0cm}
>{\centering\arraybackslash}p{3.0cm}
>{\centering\arraybackslash}X
@{}
}
\toprule

\textbf{Group} &
\textbf{TTR} &
\textbf{Avg. Sentence Length} &
\textbf{Flesch Reading Ease} &
\textbf{Flesch--Kincaid Grade Level} \\

\midrule

COVID--19\_FNIR
& 0.89
& 18.55
& 30.26
& 12.95 \\

Monkeypox
& 0.92
& 15.55
& 55.73
& 8.90 \\

\bottomrule
\end{tabularx}

\vspace{2pt}

\begin{minipage}{0.97\textwidth}
\footnotesize
\raggedright
\noindent
\textit{Note.}
TTR denotes type--token ratio (lexical diversity). Flesch Reading Ease is higher for simpler text, while Flesch--Kincaid Grade Level is higher for more complex text. Constraint is omitted here because this table focuses on the two corpora used for the cross-pandemic linguistic comparison in this section.
\end{minipage}

\end{table*}
\textbf{Engagement alignment} was evident in the VP module’s prioritisation patterns. Clusters elevated under the VP signal, as reflected by high cluster-level virality $v_j$ and, secondarily, higher viral density $V(C_j)$, consistently overlapped with those exhibiting strong emotional charge and simpler language. For example, a VP-elevated mpox cluster centred on vaccine rollout discourse combined heightened affect with readability suited for broad audiences, placing it among the most prominent narratives under the observed engagement signal despite modest overall volume. Similar patterns were observed in COVID--19\_FNIR and Constraint, where polarised or emotive framing increased cluster prominence under the proxy prioritisation signal $\tilde{E}_i$ (derived from the propensity model) rather than realised engagement. These alignments are consistent with the VP summaries in Section~\ref{sec:vp_results} and Table~\ref{tab:vp_counts}, alongside the emotion and readability evidence in Figure~\ref{fig:emotion_scores} and Table~\ref{tab:linguistic_metrics}.

\subsection{Virality Prioritisation Results}
\label{sec:vp_results}

For Monkeypox dataset, where engagement metadata are available, VP uses the observed engagement signal $E_i$ defined in Equation~\ref{eq:engagement_observed}. For COVID--19\_FNIR and Constraint, native engagement metadata are unavailable, so VP operates on the logistic regression propensity-to-spread score $\hat{p}_i$ defined in Step~\ref{step:propensity_proxy}. In this setting, \emph{viral} tweets are defined as those in the top $x \in \{1,5,10\}$ percentiles of the $\hat{p}_i$ (equivalently $\eta_i$) distribution. Cluster prioritisation is then based on the cluster-level aggregate $v_j$ (Equation~\ref{eq:cluster_virality}) using the same within-dataset percentile rule, while the viral density $V(C_j)$ (Equation~\ref{eq:viral_density}) is reported as an auxiliary diagnostic of whether high propensity is broadly distributed across a cluster or concentrated in a small subset of posts. For COVID--19\_FNIR and Constraint, these thresholds therefore reflect \emph{predicted} engagement propensity rather than observed platform interaction counts. Accordingly, we interpret VP outputs on COVID--19\_FNIR and Constraint as \emph{estimated diffusion propensity} and restrict claims to within-dataset prioritisation behaviour, while mpox VP results reflect observed interaction counts via $E_i$.

For Constraint, we therefore report VP qualitatively (case studies) rather than as definitive prevalence counts (Section~\ref{sec:constraint_instability}).

\noindent\textit{Comparability rule.}
All cross-dataset comparisons in this paper are qualitative (theme content and framing) and percentile-based (within-dataset VP ranks). We avoid interpreting differences in absolute virality magnitude across datasets, because $\tilde{E}_i$ is observed engagement for mpox but a transferred propensity logit for COVID--19\_FNIR and Constraint.

\begin{table}[htbp]
\centering
\caption{VP-flagged cluster counts under different virality thresholds.}
\label{tab:vp_counts}

\small
\begin{tabular*}{\columnwidth}{@{\extracolsep{\fill}}lccc}
\toprule
Dataset & Top 1\% & Top 5\% & Top 10\% \\
\midrule
COVID--19\_FNIR & 9/112 & 15/112 & 26/112 \\
Monkeypox      & 5/87  & 11/87  & 16/87  \\
\bottomrule
\end{tabular*}

\vspace{6pt}
\vspace{6pt}
\parbox{\columnwidth}{\footnotesize\raggedright\textit{Note (Table~\ref{tab:vp_counts}):}
``VP-flagged'' counts indicate the number of non-outlier clusters that contain at least one VP-flagged tweet (equivalently, $V(C_j)>0$), where VP-flagged tweets are those in the top-$x\%$ under the dataset-specific prioritisation signal $\tilde{E}_i$. For Monkeypox dataset, $\tilde{E}_i=E_i$ uses observed engagement. For COVID--19\_FNIR, $\tilde{E}_i=\eta_i$ uses the proxy prioritisation logit and should be interpreted as estimated diffusion propensity rather than realised engagement.}
\end{table}

Table~\ref{tab:vp_counts} shows a steady increase in VP-flagged clusters as the threshold relaxes. This pattern is expected because broader virality definitions label more tweets as viral, increasing the viral density $V(C_j)$ within clusters and consequently shifting more clusters towards the upper tail of the viral-density distribution, making high-propensity narratives more prominent under cluster-level summaries.
For COVID--19\_FNIR, more clusters achieved VP status as thresholds relaxed, reflecting a broader concentration of high-propensity content under the proxy-based virality definition. For Mpox, fewer clusters were VP-flagged under the strictest setting, but the flagged share increased as thresholds relaxed, which suggests that high-impact narratives are present but concentrated in fewer topics. This pattern is consistent with the coherence results reported in Table~\ref{tab:coherence_bertopic}, where mpox exhibited stronger internal narrative consistency.

To complement the threshold-based VP summaries, we also examine cross-dataset \emph{semantic relatedness} between representative topics. Figure~\ref{fig:dendrogram} presents a hierarchical clustering dendrogram of the top-$10$ topic representations from each dataset, based on cosine distance between topic vectors (computed from the BERTopic representations for a single fixed run (fixed seed)). This analysis is qualitative and does not depend on stable flagged-cluster counts. Accordingly, Constraint dataset can be included here to illustrate how its prominent topics relate semantically to COVID--19\_FNIR and Monkeypox datasets, even though Constraint dataset is omitted from Table~\ref{tab:vp_counts} where instability would undermine threshold-based prevalence reporting. Nevertheless, Constraint remains informative for \emph{qualitative} inspection. We emphasise that Constraint cluster examples are drawn from a single fixed run (with a fixed random seed and identical preprocessing) to illustrate the content and risk characteristics of high-propensity themes under the proxy-based scoring regime, rather than to estimate VP prevalence under each percentile threshold. Accordingly, Table~\ref{tab:vp_results} reports representative VP-flagged clusters as illustrative case studies, and these examples should not be interpreted as a complete quantitative summary of VP coverage in Constraint.

\subsubsection{Constraint clustering sensitivity}
\label{sec:constraint_instability}

Constraint is omitted from Table~\ref{tab:vp_counts} because threshold-based VP prevalence depends directly on cluster boundaries. Because VP prevalence totals depend on cluster boundaries, we treat Constraint VP prevalence as sensitivity-prone and use it for qualitative case studies rather than definitive prevalence counts.
A rerun-sensitivity check is summarised in Table~\ref{tab:constraint_stability}.

\begin{table}[!t]
\centering
\caption{BERTopic rerun-sensitivity summary for Constraint (varying UMAP seed only, $R=10$).}
\label{tab:constraint_stability}
\footnotesize
\setlength{\tabcolsep}{4pt}
\renewcommand{\arraystretch}{1.05}

\begin{tabular}{lcccc}
\toprule
$\mathbf{R}$ & \textbf{Mean ARI} & \textbf{ARI range} & $\mathbf{k}$ range & \textbf{Noise range} \\
\midrule
10 & 0.709 & [0.654, 0.770] & [42, 48] & [0.368, 0.398] \\
\bottomrule
\end{tabular}

\vspace{2pt}
\begin{flushleft}
\footnotesize\textit{Note (Table~\ref{tab:constraint_stability}):}
Mean ARI is the mean pairwise adjusted Rand index across reruns, $k$ denotes the number of non-outlier topics, and noise is the fraction of posts assigned to HDBSCAN noise ($-1$).
Only the UMAP random seed was varied.
The rerun-sensitivity check was computed under the same modelling pipeline but on a tractable rerun setting for sensitivity testing, so the reported $k$ range is not intended to match the final non-outlier topic count reported in Table~\ref{tab:clustering_results}.
\end{flushleft}
\end{table}

\noindent\textbf{Stability check (BERTopic rerun sensitivity).}
Table~\ref{tab:constraint_stability} reports a rerun-sensitivity check over $R=10$ runs in which only the UMAP random seed was varied.
The results indicate moderate partition variability alongside a high HDBSCAN noise rate, which is sufficient to make percentile-based VP prevalence totals sensitive to small boundary shifts.
Because VP prevalence totals depend directly on cluster boundaries and on which posts are assigned to non-outlier topics, we treat threshold-based VP prevalence for Constraint as sensitivity-prone and therefore focus on qualitative VP case studies rather than reporting a single definitive set of prevalence counts.

\begin{figure*}[htbp]
\centering
\includegraphics[width=\textwidth]{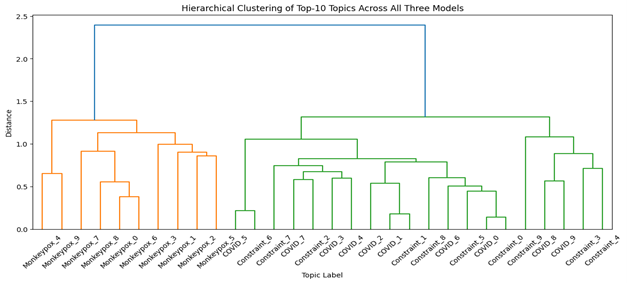}
\caption{Hierarchical clustering dendrogram of representative topic clusters identified via BERTopic-VP across COVID--19\_FNIR, Constraint, and Monkeypox datasets. Topic labels follow the convention \texttt{COVID\_k}, \texttt{Constraint\_k}, and \texttt{Monkeypox\_k}, where $k$ denotes the topic index. The vertical axis indicates semantic linkage distance under cosine-based hierarchical clustering.}
\label{fig:dendrogram}
\end{figure*}

Accordingly, we restrict quantitative threshold-based VP summaries to COVID--19\_FNIR and Monkeypox datasets and use Constraint dataset primarily for qualitative inspection (Table~\ref{tab:vp_results} and Fig.~\ref{fig:dendrogram}).

High mean coherence can coexist with \emph{run-to-run partition variability} in density-based clustering because small embedding perturbations can shift borderline points between nearby dense regions or into noise without materially changing the dominant lexical cues of the resulting topics.

\subsubsection{Qualitative Case Studies of VP-Flagged Clusters}
To illustrate the qualitative nature of VP-flagged clusters, Table~\ref{tab:vp_results} lists examples from each dataset, including virality scores, dominant terms, and misinformation presence. COVID--19\_FNIR's most prioritised cluster centred on prevention-related discourse (for example masks and cures) but contained mostly factual examples in the sampled items. In contrast, the Monkeypox dataset contained some of the highest-scoring clusters, driven by credible news narratives about response and case trends. In the examined run, the Constraint case-study clusters contained a higher share of misinformation-labelled examples, so we treat them as illustrative high-priority narratives rather than a population-level estimate of VP prevalence or amplification risk.

\begin{table*}[!t]
\centering
\caption{Representative VP-flagged clusters with cluster-level virality score $v_j$.}
\label{tab:vp_results}

\small
\setlength{\tabcolsep}{5pt}
\renewcommand{\arraystretch}{1.15}

\begin{tabularx}{\textwidth}{
@{}
>{\raggedright\arraybackslash}p{2.6cm}
>{\centering\arraybackslash}p{1.6cm}
>{\centering\arraybackslash}p{3.0cm}
>{\raggedright\arraybackslash}X
>{\raggedright\arraybackslash}p{2.8cm}
@{}
}
\toprule

\textbf{Dataset} &
\textbf{Cluster ID} &
\textbf{Virality Score (\(v_j\))} &
\textbf{Top Terms} &
\textbf{Misinfo Presence} \\

\midrule

COVID--19\_FNIR
& 1
& 3.457
& masks, cure, prevent
& Mostly factual \\

Monkeypox
& 11
& \textbf{13.000}
& scientists, losing fight
& None detected \\

Monkeypox
& 25
& 9.765
& cases, states, prevalence
& None detected \\

Constraint
& 2
& 2.681
& lagos, discharged, fct
& All misinformation \\

Constraint
& 4
& 2.649
& covid, cases, reported
& Mixed \\

\bottomrule
\end{tabularx}

\vspace{2pt}

\begin{minipage}{0.97\textwidth}
\footnotesize
\raggedright
\noindent
\textit{Note.}
For Monkeypox, \(v_j\) is the mean observed engagement \(E_i\) over tweets in cluster \(j\).
For COVID--19\_FNIR and Constraint, \(v_j\) is the mean proxy score over tweets in cluster \(j\), computed from the logistic propensity model, using the logit score \(\eta_i\), with \(\hat{p}_i=\sigma(\eta_i)\).
Accordingly, VP rankings for COVID--19\_FNIR and Constraint reflect estimated diffusion propensity, not realised platform engagement.
\end{minipage}

\end{table*}

\subsubsection{Synthesis of Findings Across Detection, Topics, and Virality}
Across datasets, BERTopic-VP separated narratives effectively and supported accurate supervised detection. Monkeypox dataset yielded the best clustering quality and strongest classifier performance, suggesting clearer narrative boundaries and more consistent lexical signals. COVID--19\_FNIR’s lower coherence reflected its diverse themes, including media reuse, prevention claims, and geographic counts. Constraint produced coherent clusters but with weaker label alignment, reflected in lower purity and ARI.

Importantly, the VP-flagged clusters in each dataset often overlapped with high classifier confidence regions, as measured by F1-score and ROC-AUC (Section~\ref{subsec:supervised}). This confirms that topics prioritised for virality were also those where the misinformation classifier achieved high discriminatory power, indicating a strong intersection between engagement-driven spread and detectable linguistic-rhetorical patterns.

Virality analysis revealed that COVID--19\_FNIR and Monkeypox datasets both contained highly viral yet largely factual clusters, while Constraint’s flagged clusters contained mostly misinformation. This demonstrates that VP is sensitive to engagement concentration regardless of veracity, which reinforces the importance of coupling VP with the misinformation classifier. In practice, this combined signal enables analysts to prioritise clusters that are both highly viral and likely false, thereby identifying emerging threats with the greatest potential for rapid diffusion and harm.

When VP outputs are linked to emotional-linguistic drivers (Table~\ref{tab:sentiment_emotion}, Table~\ref{tab:linguistic_metrics}) and coherence patterns (Table~\ref{tab:coherence_bertopic}), we establish a multi-dimensional validation of high-risk misinformation topics. The VP mechanism consistently surfaces narratives with both strong engagement potential and thematic alignment, supporting its role as a proactive tool for public health counter-messaging. Future work could integrate rhetorical stance analysis to further refine prioritisation beyond engagement metrics alone.

\subsection{Coherence as Supporting Validation}
\label{subsec:coherence}
The VP results (Section~\ref{sec:vp_results}) identified clusters with high virality-signal density (observed engagement in the Monkeypox dataset and the proxy VP score in COVID--19\_FNIR and Constraint), many of which corresponded to thematically cohesive narratives.

To assess whether semantic structure supported these virality patterns, we computed topic coherence scores for all datasets. While emotional tone, lexical simplicity, and engagement potential were the dominant drivers of VP prioritisation, topic coherence provided secondary validation of these findings (Table~\ref{tab:coherence_bertopic}). Mpox misinformation topics achieved a higher average coherence score (\textbf{0.286}) than COVID--19\_FNIR topics (\textbf{0.084}), with Constraint topics reaching \textbf{0.463}. The higher scores for Monkeypox and Constraint datasets suggest a tighter thematic focus and internally consistent framing compared to the more fragmented COVID--19\_FNIR narratives.

This concentrated narrative structure, when coupled with emotionally charged and easily digestible language, further increased the likelihood of amplification. Evidence for this relationship is provided in Table~\ref{tab:sentiment_emotion}, which shows that mpox misinformation exhibited the most intense emotional activation, particularly for fear (\textbf{0.226}) and anger (\textbf{0.193}). Table~\ref{tab:linguistic_metrics} complements this by demonstrating that mpox tweets were also linguistically simpler, with higher Flesch Reading Ease scores and lower Flesch--Kincaid Grade Levels, reducing cognitive barriers to engagement.

In contrast, COVID--19\_FNIR misinformation, though less coherent according to Table~\ref{tab:clustering_results}, still produced high-impact clusters when emotional salience and lexical accessibility aligned, for example, the \textit{``Italy coffin''} shock imagery narrative, which leveraged strong visual cues alongside emotionally charged text. This interplay between emotional intensity, low cognitive load, and high engagement signals reinforces the conclusion that while coherence supports interpretability, it is the psychological and stylistic dimensions that most directly drive virality in health misinformation.

Within the Monkeypox dataset (mean $C_v = 0.286$, Table~\ref{tab:clustering_results}), coherence varies across clusters, indicating the presence of both more fragmented discussions and more cohesive narrative themes. To illustrate these contrasts, Table~\ref{tab:coherence_examples} presents paraphrased examples from representative low- and high-coherence clusters, including mpox ($C_v=0.22$ vs.\ $0.33$) and a high-coherence Constraint example ($C_v=0.463$). Collectively, these examples highlight the distinction between fragmented, multi-theme discussions and tightly focused frames that exhibit greater internal consistency.

Low coherence in COVID--19\_FNIR reflects topic fragmentation rather than model failure. Qualitative inspection suggested that some clusters combine lab-leak, mask-efficacy, and policy narratives, which reduces within-topic consistency. As an analyst-facing heuristic for short-form Twitter data, one could apply a minimum coherence threshold (for example $C_v < 0.10$) and optionally merge low-coherence clusters using post hoc centroid similarity (cosine $> 0.65$). This heuristic is not applied to the reported results in this paper.

\begin{table*}[!t]
\centering
\caption{Qualitative examples of low- vs.\ high-coherence clusters in COVID--19\_FNIR, Constraint and Monkeypox datasets}
\label{tab:coherence_examples}
\footnotesize
\setlength{\tabcolsep}{4pt}
\renewcommand{\arraystretch}{1.15}

\begin{tabularx}{\textwidth}{
  >{\raggedright\arraybackslash}p{2.3cm}
  >{\raggedright\arraybackslash}p{3.2cm}
  >{\raggedright\arraybackslash}X
  >{\centering\arraybackslash}p{1.4cm}
}
\toprule
\textbf{Dataset} & \textbf{Cluster theme} & \textbf{Representative tweets (paraphrased)} & $\mathbf{C}_v$ \\
\midrule
COVID--19\_FNIR (Low) & Mixed preventive and political claims &
\enquote{The vaccine is useless, the virus is fake, and the government just wants control.}
\par\enquote{Masks work but lockdowns destroy jobs.} The framing conflicts across posts. & 0.084 \\

Constraint (High) & Vaccine side-effects narrative &
\enquote{My friend took the shot and got sick for weeks.}
\par\enquote{Vaccines have dangerous chemicals that doctors hide.} A single, consistent misinformation frame. & 0.463 \\

Mpox (Low) & Broad outbreak speculation &
\enquote{This virus comes from labs in Europe.}
\par\enquote{It is fake news made up by the media.} Mixed conspiracy and denial themes. & 0.22 \\

Mpox (High) & Vaccine rollout critique &
\enquote{People in rural areas still cannot access vaccines.}
\par\enquote{Authorities promise doses but none arrive.} Focused grievance on distribution failure. & 0.33 \\
\bottomrule
\end{tabularx}
\begin{flushleft}
\footnotesize\textit{Note (Table~\ref{tab:coherence_examples}):} The representative qualitative examples have been paraphrased to minimise identifiability and to support ethical reporting, values for low/high clusters are relative to each dataset's mean $C_v$.
\end{flushleft}
\end{table*}

\subsubsection{Justification for Coherence Score Differences in BERTopic-VP}
To quantify semantic coherence, we computed the mean topic coherence ($C_v$) for each dataset processed through BERTopic-VP.

\noindent\textbf{Bootstrap test for mean coherence.}
For each dataset, we fixed a single BERTopic run (fixed preprocessing and random seed) and computed per-topic coherence values $\{C_{v,j}\}_{j=1}^{k}$ over non-outlier topics.
We then performed $B=10{,}000$ bootstrap resamples of topics (sampling $\{C_{v,j}\}$ with replacement) to obtain a bootstrap distribution for the mean coherence $\bar{C}_v$ and for pairwise differences $\Delta=\bar{C}_v^{(A)}-\bar{C}_v^{(B)}$ between datasets.
This procedure treats the topic partition as fixed and quantifies uncertainty in $\bar{C}_v$ arising from topic-to-topic variability rather than from re-estimating the clustering model.

To assess whether mean coherence differed between datasets, we used a two-sided bootstrap test based on repeated resampling of topic-level coherence values. This evaluates the null hypothesis that the observed difference in mean coherence could arise by chance in either direction. The test indicated a statistically significant difference between COVID--19\_FNIR and Monkeypox datasets mean coherence ($p<0.05$).

Percentile bootstrap 95\% intervals for $\bar{C}_v$ were $[0.076,\,0.091]$ for COVID--19\_FNIR, $[0.273,\,0.299]$ for Mpox, and $[0.451,\,0.475]$ for Constraint (Table~\ref{tab:coherence_bertopic}).

\begin{table*}[!t]
\centering
\caption{Within-study BERTopic-VP coherence and externally reported BERTopic coherence values used for contextual comparison.}
\label{tab:coherence_bertopic}

\small
\setlength{\tabcolsep}{5pt}
\renewcommand{\arraystretch}{1.15}

\begin{tabularx}{\textwidth}{
@{}
>{\raggedright\arraybackslash}p{3.8cm}
>{\raggedright\arraybackslash}p{4.2cm}
>{\centering\arraybackslash}p{1.8cm}
>{\raggedright\arraybackslash}X
@{}
}
\toprule

\textbf{Model / Study} &
\textbf{Dataset / Domain} &
$\mathbf{C}_v$ &
\textbf{Notes} \\

\midrule

\textbf{BERTopic-VP (this study)}
& COVID--19\_FNIR misinformation tweets
& 0.084
& Lowest coherence, reflecting narrative fragmentation. \\

\textbf{BERTopic-VP (this study)}
& Mpox misinformation tweets
& \textbf{0.286}
& Higher cohesion, with fewer dominant frames. \\

\textbf{BERTopic-VP (this study)}
& Constraint misinformation tweets
& 0.463
& Highest cohesion in this study's datasets. \\

BERTopic \textbf{(reported)} \cite{yin2022sentiment}
& COVID--19 vaccine tweets
& 0.52
& Higher coherence due to a more thematically uniform dataset. \\

BERTopic \textbf{(reported)} \cite{amara2021multilingual}
& Facebook COVID--19 posts
& 0.53
& Performs comparably to and, in some settings, modestly better than LDA on multilingual health data. \\

\bottomrule
\end{tabularx}

\vspace{2pt}

\begin{minipage}{0.97\textwidth}
\footnotesize
\raggedright
\noindent
\textit{Note.}
Rows 1--3 report coherence computed in this study under a single fixed preprocessing and \(C_v\) protocol applied to the study datasets. Rows 4--5 reproduce \(C_v\) values reported in the cited papers and are included for contextual comparison only. These external values were not re-implemented on the study corpora and should not be interpreted as a head-to-head benchmark because datasets, preprocessing procedures, topic granularity, and coherence implementations differ across studies.
\end{minipage}

\end{table*}

The disparity is primarily explained by narrative fragmentation in COVID--19\_FNIR, where topics encompassed diverse and competing themes such as lab-leak theories, vaccine-efficacy debates, anti-lockdown rhetoric, and statistical scepticism. In contrast, mpox discourse was anchored in a smaller number of frames, most notably institutional trust, vaccine rollout criticism, and outbreak severity, yielding higher cluster cohesion. Constraint, being task-focused and thematically narrower, produced the highest coherence in our datasets. Importantly, these differences were observed under the same fixed UMAP and HDBSCAN configuration applied across datasets, without dataset-specific retuning.

Comparisons with existing literature (Table~\ref{tab:coherence_baselines}) further contextualise these results. Table~\ref{tab:coherence_baselines} reports coherence values as stated in the cited studies, and these are not recomputed on our datasets. Reported LDA coherence values on short-text pandemic corpora vary widely, for example around $\sim$0.47 for Twitter vaccine discourse \cite{egger2022topic} and as low as 0.031 for multi-domain comment data \cite{de2022experiments}. These reference points are not a direct like-for-like benchmark against our experiments because datasets, preprocessing choices, and coherence implementations differ across studies. They are therefore used only to situate our observed coherence levels within the broader short-text topic-modelling literature, rather than to claim strict superiority.

\begin{table*}[!t]
\centering
\caption{Selected reported \(C_v\) coherence values for a classical baseline (LDA) on short-text pandemic corpora, included for context only and not re-implemented in this paper.}
\label{tab:coherence_baselines}

\small
\setlength{\tabcolsep}{6pt}
\renewcommand{\arraystretch}{1.15}

\begin{tabularx}{\textwidth}{
@{}
>{\raggedright\arraybackslash}p{3.5cm}
>{\raggedright\arraybackslash}p{4.0cm}
>{\centering\arraybackslash}p{2.0cm}
>{\raggedright\arraybackslash}X
@{}
}
\toprule

\textbf{Model / Study} &
\textbf{Dataset / Domain} &
$\mathbf{C}_v$ &
\textbf{Notes} \\

\midrule

LDA \cite{egger2022topic}
& Twitter vaccine discourse
& \(\sim 0.47\)
& Lower robustness in microtext settings. \\

LDA \cite{de2022experiments}
& Multi-domain comments
& 0.031
& Lowest performance on short-text corpora. \\

\bottomrule
\end{tabularx}

\vspace{2pt}

\begin{minipage}{0.97\textwidth}
\footnotesize
\raggedright
\noindent
\textit{Note.}
The \(C_v\) values shown are reproduced from the cited studies and are included for contextual comparison only. They are not recomputed on COVID--19\_FNIR, Mpox, or Constraint in this paper.
\end{minipage}

\end{table*}

\subsection{Robustness checks and descriptive validation}
\label{subsec:stat_validation}
To assess whether the inter-dataset contrasts reported in Section~\ref{sec:results} were stable under alternative summaries, we carried out a set of \emph{informal robustness checks}. These checks are reported descriptively and are not treated as definitive hypothesis tests, because some of the units of analysis (for example, cross-validation folds and model-derived topics) are not independent observations.

For classification performance, five-fold cross-validation scores on the \emph{training split} were used to summarise within-dataset variability in macro-$F_1$. Fold-level values were not treated as independent samples for formal inference; instead, we inspected overlap in fold-to-fold performance ranges across datasets. This inspection indicated a narrow and largely overlapping macro-$F_1$ band (approximately $0.91$--$0.95$), consistent with the small between-dataset differences observed in the headline results reported on the held-out test split (Table~\ref{tab:classification_results}).

For topic quality, coherence was computed at the topic level (one $C_v$ value per topic). Because topics are generated within a shared modelling pipeline and may not be statistically independent, we report the comparison of mean coherence between COVID--19\_FNIR and Monkeypox datasets as a descriptive contrast, supported by the direction and magnitude of the observed difference (Table~\ref{tab:coherence_bertopic}), rather than as a strict inferential claim.

For affective language, emotion scores were computed at the tweet level and contrasted across datasets by inspecting the direction and magnitude of mean differences across emotion categories (Table~\ref{tab:sentiment_emotion}). Because tweets within the same conversation, community, or time window may be correlated, we report these contrasts descriptively rather than as definitive hypothesis tests. The results indicate that fear and anger differences persist under alternative summaries and are not driven by a single emotion category, while remaining conditional on dataset provenance and measurement choices.

These robustness checks assess sensitivity to key analytical choices, and the results are consistent with the main pattern of inter-dataset contrasts, although they do not eliminate alternative explanations linked to corpus construction and platform context.

\section{Discussion}
This paper examined how health misinformation narratives differ across COVID--19\_FNIR, Monkeypox, and Constraint datasets, and assessed whether the proposed \textsc{BERTopic-VP} framework can support both accurate detection and operationally useful prioritisation. Overall, the results show that combining semantic topic discovery with a virality-aware prioritisation layer yields a practical early-warning layer, while the supervised component maintains strong discrimination between misinformation and non-misinformation content.

From a classification perspective, the supervised misinformation classifier within the \textsc{BERTopic-VP} pipeline achieved consistently strong performance across all datasets, with F1-scores reaching up to \textbf{0.950} and ROC--AUC values up to \textbf{0.989} (Table~\ref{tab:classification_results}). This indicates that misinformation remains separable even under heterogeneous short-text conditions, and suggests that the semantic representations and engineered cues used downstream capture stable signals of veracity across domains. In practical terms, this supports deployment as a screening layer, particularly when analyst capacity is limited and false positives carry operational cost.

The linguistic results highlight that misinformation does not adopt a uniform style across health crises. COVID--19\_FNIR exhibited greater thematic fragmentation and higher complexity of lexical and readability (Table~\ref{tab:linguistic_metrics}), which can plausibly increase perceived credibility through an ``expert-like'' register, consistent with accounts of how analytic presentation can reduce scrutiny in fast-scrolling environments \cite{pennycook2019fighting}. By contrast, mpox misinformation was comparatively simpler in surface form but more affectively intense, with narratives often framed through politicised, alarming, or identity-linked language. Constraint showed an intermediate profile, combining a relatively clear lexical style with stronger polarisation in stance and rhetorical tone, which is consistent with its hostile or adversarial discourse framing.

Emotional tone analysis provides a behavioural interpretation of why certain topics are more likely to attract interaction. Figure~\ref{fig:emotion_scores} contrasts COVID--19 and mpox emotion distributions and shows stronger fear- and anger-linked activation for mpox relative to COVID--19\_FNIR, which aligns with the broader claim that high-arousal emotions increase sharing and moralised engagement \cite{brady2017emotion}. Importantly, the emotional results should be read alongside the cluster-level prioritisation outputs rather than in isolation, because VP is designed to surface topics where such affective cues coincide with concentrated virality signals.

Topic coherence provides secondary evidence about how consolidated each narrative space is. Monkeypox dataset achieved higher coherence than COVID--19\_FNIR (Table~\ref{tab:coherence_bertopic}), suggesting tighter thematic focus and more internally consistent framing, whereas COVID--19\_FNIR reflects a broader and noisier mixture of narratives. Constraint achieved the highest coherence, which indicates strong within-topic semantic consistency, although its weaker alignment with labels (for example lower ARI) implies that coherent topics are not necessarily cleanly separated by the dataset's veracity annotations. These patterns are useful for interpretation, but coherence is not treated as the primary success criterion in this work because the operational goal is early warning and prioritisation rather than perfect topic purity.

The VP results reinforce the practical value of separating \emph{topic discovery} from \emph{risk prioritisation}. For Mpox, VP operates on \emph{observed} engagement metadata, so ``viral'' tweets are defined using the empirical engagement distribution and cluster flagging reflects genuine platform interaction concentrations. For COVID--19\_FNIR and Constraint, engagement metadata are unavailable, so VP instead uses the logistic regression propensity-to-spread scores, meaning that ``viral'' is defined relative to the \emph{predicted} propensity distribution rather than observed retweet or like counts. This distinction matters for interpretation: within-dataset prioritisation remains meaningful, but cross-dataset comparisons between mpox (observed engagement) and COVID--19\_FNIR or Constraint (proxy engagement) should be treated cautiously, because the underlying virality signals differ in what they measure. For COVID--19\_FNIR and Constraint, VP uses the logistic regression propensity-to-spread score $\hat{p}_i$ (Step~\ref{step:propensity_proxy}), so ``viral'' is defined relative to predicted propensity rather than observed likes or retweets. 

Although median \emph{observed} engagement in Monkeypox dataset and median \emph{proxy} propensity-to-spread in COVID--19\_FNIR dataset were low, both corpora contained high-impact outliers with the potential for disproportionate diffusion. Under the top 1\% threshold, 9 of 112 clusters (8.0\%) were VP-flagged in COVID--19\_FNIR, while 5 of 87 clusters (5.7\%) were VP-flagged in Monkeypox dataset (Table~\ref{tab:vp_counts}). As thresholds relaxed, VP-flagged cluster counts increased in both datasets, which is expected because broader percentile cut-offs label more tweets as viral and raise cluster-level summary scores (both $v_j$ and the auxiliary viral density $V(C_j)$). Together, these results support VP as an early-warning mechanism that can surface low-volume but high-risk narratives for analyst review, particularly when coupled with the veracity classifier.

Within these constraints, VP still provides actionable output. The representative clusters in Table~\ref{tab:vp_results} illustrate that VP can surface high-impact narratives even when overall tweet volume is modest. The Constraint examples are particularly informative: the sampled VP-flagged clusters include cases marked as predominantly misinformation, suggesting that in this corpus the proxy-based virality signal can coincide with higher-risk content. At the same time, the presence of highly prioritised but largely factual clusters in COVID--19\_FNIR and Monkeypox datasets reinforces a key operational point, VP is intentionally veracity-agnostic, so it should be coupled with the misinformation classifier to prioritise clusters that are \emph{both} high-impact and likely false.

The hierarchical dendrogram (Figure~\ref{fig:dendrogram}) complements the VP summaries by providing a qualitative view of how topic representations relate across datasets under cosine-based linkage. This analysis is useful for assessing whether narratives cluster by outbreak domain or show thematic overlap across crises. In particular, the dendrogram suggests partial separation between Mpox-related themes and COVID--19 or Constraint themes, supporting the claim that narrative content and framing are outbreak-specific and that unsupervised discovery is necessary to detect emergent topics without relying on static keyword lists. High lexical diversity values (TTR $>$ 0.89) across datasets should also be interpreted carefully, because short-text contexts can inflate diversity without necessarily indicating deeper semantic complexity \cite{mccarthy2010mtld}.

\subsubsection{Implications for infodemic surveillance.}
Taken together, the results support \textsc{BERTopic-VP} as a surveillance-oriented pipeline that links semantic clustering, affective and linguistic diagnostics, and a practical prioritisation mechanism. Its primary operational value lies in structuring analyst attention, by surfacing clusters that exhibit early signals of high diffusion risk or persuasive traction, even when topic volume is modest. In a public health context, this supports triage workflows in which analysts can first review high-priority clusters, and then apply misinformation detection and domain verification to separate harmful claims from high-engagement but factual reporting. The practical implications should be interpreted in light of the data and measurement constraints discussed in Section~\ref{sec:limitations}.

To complement the performance and descriptive findings reported above, the following subsection formalises how \textsc{BERTopic-VP} supports interpretability and explainability through explicit topic representations, transparent virality scoring, and observable linguistic and engagement diagnostics.

\subsection{Explainability \& Interpretability of the BERTopic-VP Pipeline}
\label{sec:bertopic_vp_interpretability}
BERTopic-VP was designed to support interpretable and explainable analysis alongside strong performance, through a modular pipeline in which semantic representation, dimensionality reduction, clustering, and virality scoring are separated and can be inspected independently (Section~\ref{sec:materials_methods}, Figure~\ref{fig:end_to_end_vp_pipeline}). Topics are formed as tweet clusters and represented using c\textsc{tf}--idf keywords and representative examples, enabling each cluster to be read as a concrete narrative rather than an opaque latent factor. In practice, exemplars are selected as high-salience posts within each cluster (for example, those closest to the cluster centre in embedding space or those with high within-cluster relevance under the topic representation). The clustering outputs in Table~\ref{tab:clustering_results} and the coherence analyses in Sections~\ref{subsec:coherence} and \ref{subsec:stat_validation} provide quantitative support for the topic structure, while the virality prioritisation stage allows topic- and tweet-level outputs to be justified using observable linguistic signals and, where available, observed engagement counts or otherwise the proxy prioritisation score used to compute the prioritisation signal.

Explainability is provided by the VP overlay and the explicit aggregation of the prioritisation signal into cluster-level scores. As described in Section~\ref{sec:vp_results}, BERTopic-VP computes a tweet-level prioritisation signal $\tilde{E}_i$, using observed engagement $E_i$ for Monkeypox dataset and a transferred propensity proxy (logit $\eta_i$) for COVID--19\_FNIR and Constraint (Equations~\ref{eq:engagement_observed} and \ref{eq:engagement_surrogate}). Cluster-level virality is then obtained by aggregating $\tilde{E}_i$ into $v_j$ (Equation~\ref{eq:cluster_virality}), and prioritisation is operationalised using within-dataset percentile thresholds $x \in \{1,5,10\}$ applied to $\tilde{E}_i$ and $v_j$. For additional cluster screening, viral concentration is summarised by the viral density score $V(C_j)$ (Equation~\ref{eq:viral_density}), which highlights clusters where a small subset of highly ranked posts drives disproportionate spread potential.

At the \textbf{topic level}, explainability is strengthened by linking VP status to emotion, readability, and coherence diagnostics. Sections~\ref{sec:emotional_lexical_engagement} and \ref{subsec:coherence} show that VP-flagged clusters often combine strong emotional activation (Table~\ref{tab:sentiment_emotion}, Figure~\ref{fig:emotion_scores}), relatively simple language (Table~\ref{tab:linguistic_metrics}), and moderate to high coherence (Table~\ref{tab:coherence_bertopic}). This supports a compact, stakeholder-facing account of risk, for example, a coherent mpox narrative that is fear-laden, lexically simple, and unusually prominent under the VP signal, linking pipeline outputs to behavioural mechanisms discussed earlier in this paper.

At the \textbf{tweet level}, local explanations are supported by connecting classifier outputs to topic membership and virality context. When a tweet is classified as likely misinformation by the supervised component (Table~\ref{tab:classification_results}), BERTopic-VP can contextualise that output via the tweet's topic cluster, the VP status of that cluster, and the observable linguistic cues used by the prioritisation signal (Section~\ref{sec:emotional_lexical_engagement}). This layered pathway from tweet to topic to VP status provides a more actionable explanation than a probability score alone, consistent with expectations for explainable systems in high-stakes public health monitoring.

Interpretability is further supported by cross-dataset comparison. The dendrogram in Figure~\ref{fig:dendrogram} provides an interpretable map of semantic proximity between COVID--19 and mpox narratives, while differences in coherence and emotion profiles across datasets (Tables~\ref{tab:coherence_bertopic} and \ref{tab:sentiment_emotion}) help explain why some corpora yield tightly focused but polarised themes and others exhibit greater fragmentation. These contrasts enable analysts to understand not only which clusters are high-priority, but also how narrative structure and emotional framing contribute to virality risk.

Taken together, BERTopic-VP provides explainability through explicit topic representations, transparent VP flagging rules, and observable linguistic and engagement diagnostics, enabling outputs to be communicated and audited by non-technical stakeholders responsible for infodemic management.

\subsection{Limitations}
\label{sec:limitations}

This study has several limitations. First, the COVID--19 datasets (COVID--19\_FNIR and Constraint) may exclude high-engagement misinformation that was removed or never captured. The Monkeypox dataset, which was collected via keyword search, may contain irrelevant noise and does not capture misinformation circulating in closed networks or in other languages.

Second, engagement metrics were incomplete, particularly for quote tweets and replies, which constrained interaction-aware analysis and limited the interpretability of engagement pathways beyond simple aggregates. Although BERTopic-VP integrates multiple engagement types, the absence of complete quote-tweet and reply data introduces a partial visibility bias. Quote tweets often contain counter-speech or corrective commentary that differ structurally from retweets, which may lead to underestimation of dialogic virality. For COVID--19\_FNIR and Constraint, virality analysis relies on a model-based engagement propensity rather than raw platform counts, because native metadata are unavailable. This proxy captures relative susceptibility to engagement given linguistic and psychological cues, but it cannot fully substitute for observed likes or retweets. Accordingly, VP comparisons for COVID--19\_FNIR and Constraint should be interpreted in terms of within-dataset ranks and percentile thresholds on $\tilde{E}_i$, rather than absolute engagement magnitude.

We provide a preliminary rerun-sensitivity check for Constraint by varying the UMAP seed (Table~\ref{tab:constraint_stability}), which indicates that VP prevalence totals can be sensitive to small partition perturbations when the HDBSCAN noise rate is high. More extensive stability analysis, including uncertainty intervals for VP prevalence totals under different percentile thresholds and broader hyperparameter perturbations, is left for future work. Nonetheless, the qualitative narratives surfaced by VP were consistent across the inspected runs under the current signal definition.

Sentiment and emotion estimates depend on lexicon-based tools (for example TextBlob for polarity and NRC lexicons for discrete emotions), which may struggle with sarcasm, implicit emotion, or cultural variation. Discrepancies between tabulated and visualised emotion scores reflect differences in aggregation methods (mean values compared with distributional densities). These inconsistencies warrant caution when interpreting the emotional tone of misinformation narratives.

The study also does not measure belief change or the effectiveness of corrective interventions. Engagement and sentiment provide indirect indicators of audience response, but they do not capture whether exposure to misinformation alters attitudes or behaviours. Future research should incorporate behavioural outcomes, user-level propagation patterns, and the impact of counter-messaging or fact-checking.

Generalisability is further constrained by language, platform, and period. All analyses are restricted to English-language Twitter data from 2020 to 2022. Virality patterns, narrative styles, and platform affordances may differ on services such as TikTok or WeChat, where audiovisual cues and recommendation algorithms play a more prominent role, and in regions that employ different public-health framings. Multilingual extensions will require fine-tuned sentence-transformer models and culturally specific lexicons.

Although the Measure of Textual Lexical Diversity (MTLD) \cite{mccarthy2010mtld} offers a length-independent alternative to the type-token ratio (TTR), its application to short microtext such as tweets is methodologically unstable. MTLD requires longer contiguous text sequences to calculate reliable factor thresholds, whereas most tweets contain far fewer than the recommended minimum of 50 tokens. Preliminary testing revealed high variance in MTLD values due to tweet brevity, code-switching, and token noise. TTR was therefore retained to maintain comparability with prior misinformation studies and to ensure interpretive consistency across datasets. Future work could apply MTLD at the topic-cluster or user-aggregate level, where token sequences are sufficiently long for stable estimation.

\subsubsection{Adversarial robustness \& account authenticity}
\label{sec:adversarial_robustness}

Although VP helps surface high-impact narratives, it implicitly assumes that the prioritisation signal $\tilde{E}_i$ reflects genuine user attention. In practice, coordinated inauthentic behaviour, including bot networks and organised troll activity, can inflate likes and retweets and thereby bias VP rankings towards manufactured narratives, distorting perceived diffusion risk \cite{ferrara2016rise, shao2018spread}. This complicates the interpretation of virality signals and may lead analysts to overestimate the organic reach of particular topics. A related concern applies when virality is approximated via the propensity-to-spread proxy, because coordinated accounts can also shape the linguistic and interactional patterns that the proxy learns from, potentially transferring manipulation artefacts into proxy-based prioritisation.

Future work should therefore incorporate adversarial robustness by down-weighting engagement contributions from accounts with low authenticity. For example, Botometer-style classifiers estimate the probability that an account is automated, while complementary coordination indicators can flag synchronised posting or amplification \cite{davis2016botornot, varol2017online}. One simple extension is to define an authenticity weight $a_u \in [0,1]$ for author $u$ and compute an adjusted score $E_i^{(\mathrm{auth})}=a_{u(i)}\,E_i$ (and analogously adjust proxy training by excluding or down-weighting low-authenticity accounts). Authenticity-weighted prioritisation would help distinguish organic from manipulated spread and provide a more reliable indicator of genuine public attention, particularly when infodemics are influenced by strategic information operations. This extension would also improve cross-event comparability, because the same narrative may attract very different mixtures of organic and inauthentic engagement across crises and regions.

\section{Conclusion}
This study developed and evaluated \textsc{BERTopic-VP}, a modular, virality-aware topic-modelling pipeline for detecting and interpreting health misinformation within a monitoring-oriented workflow. Its design is compatible with periodic near-real-time batch refreshes, although live-streaming performance and operational scalability were not evaluated. Applied to the COVID--19\_FNIR, Constraint, and \textit{Monkeypox} datasets, the pipeline revealed clear cross-event differences in narrative structure, emotional tone, and prioritisation
patterns, with the most direct contrast observed between COVID--19\_FNIR and the \textit{Monkeypox} dataset. COVID--19 misinformation was more thematically diverse and linguistically complex, whereas mpox narratives were narrower, more affectively charged, and more frequently framed through politicised or identity-linked discourse.

Although median engagement was low where engagement was observable (Monkeypox dataset), all datasets contained a small number of high-impact outliers with the potential for disproportionate diffusion. In COVID--19\_FNIR and Constraint, the corresponding virality signal is proxy-based (propensity-to-spread) rather than observed interaction counts because native engagement metadata are unavailable. Together, these findings reinforce the practical value of coupling semantic topic discovery with a virality-aware prioritisation layer, enabling analysts to surface low-volume but high-risk narratives before they dominate attention. In contrast to frequency-based topic models, which can degrade under short-text sparsity, \textsc{BERTopic-VP} maintains interpretable clustering for microtext and supports modular deployment for public health surveillance workflows.

For public health practitioners, the framework provides a replicable mechanism for prioritising narratives that warrant rapid review and counter-messaging. For researchers, the results highlight how narrative coherence, emotional resonance, and accessibility cues can interact with diffusion potential, and why engagement-aware design is important when translating topic models into operational infodemic monitoring tools.

Future work should extend \textsc{BERTopic-VP} to multilingual settings and platform-specific contexts (for example TikTok and WhatsApp), and incorporate network and behavioural traces where available. Longitudinal evaluation is also needed to test whether early VP-flagged clusters anticipate subsequent surges in observed engagement where measurable, mainstream coverage, or offline behavioural response, thereby strengthening the model’s utility as an early-warning infodemic indicator.

\bibliographystyle{ieeetr}
\bibliography{refs}

@article{do2022infodemics,
  title={Infodemics and health misinformation: a systematic review of reviews},
  author={Do Nascimento, Israel Junior Borges and Pizarro, Ana Beatriz and Almeida, Jussara M and Azzopardi-Muscat, Natasha and Gon{\c{c}}alves, Marcos Andr{\'e} and Bj{\"o}rklund, Maria and Novillo-Ortiz, David},
  journal={Bulletin of the World Health Organization},
  volume={100},
  number={9},
  pages={544},
  year={2022},
  publisher={World Health Organization}
}

@article{di2022health,
  title={Health misinformation detection in the social web: An overview and a data science approach},
  author={Di Sotto, Stefano and Viviani, Marco},
  journal={International Journal of Environmental Research and Public Health},
  volume={19},
  number={4},
  pages={2173},
  year={2022},
  publisher={MDPI}
}

@article{suarez2021prevalence,
  title={Prevalence of health misinformation on social media: systematic review},
  author={Suarez-Lledo, Victor and Alvarez-Galvez, Javier},
  journal={Journal of medical Internet research},
  volume={23},
  number={1},
  pages={e17187},
  year={2021},
  publisher={JMIR Publications Toronto, Canada}
}

@article{luo2023exploring,
  title={Exploring the impact of sentiment on multi-dimensional information dissemination using COVID-19 data in China},
  author={Luo, Han and Meng, Xiao and Zhao, Yifei and Cai, Meng},
  journal={Computers in Human Behavior},
  volume={144},
  pages={107733},
  year={2023},
  publisher={Elsevier}
}

@article{melki2021mitigating,
  title={Mitigating infodemics: The relationship between news exposure and trust and belief in COVID-19 fake news and social media spreading},
  author={Melki, Jad and Tamim, Hani and Hadid, Dima and Makki, Maha and El Amine, Jana and Hitti, Eveline},
  journal={Plos one},
  volume={16},
  number={6},
  pages={e0252830},
  year={2021},
  publisher={Public Library of Science San Francisco, CA USA}
}

@article{muhammed2022disaster,
  title={The disaster of misinformation: a review of research in social media},
  author={Muhammed T, Sadiq and Mathew, Saji K},
  journal={International journal of data science and analytics},
  volume={13},
  number={4},
  pages={271--285},
  year={2022},
  publisher={Springer}
}

@article{ravichandran2023classification,
  title={Classification of Covid-19 misinformation on social media based on neuro-fuzzy and neural network: A systematic review},
  author={Ravichandran, Bhavani Devi and Keikhosrokiani, Pantea},
  journal={Neural Computing and Applications},
  volume={35},
  number={1},
  pages={699--717},
  year={2023},
  publisher={Springer}
}

@article{rudat2015making,
  title={Making retweeting social: The influence of content and context information on sharing news in Twitter},
  author={Rudat, Anja and Buder, J{\"u}rgen},
  journal={Computers in human behavior},
  volume={46},
  pages={75--84},
  year={2015},
  publisher={Elsevier}
}

@article{saini2022association,
  title={The association between dissemination and characteristics of pro-/anti-COVID-19 vaccine messages on Twitter: application of the elaboration likelihood model},
  author={Saini, Vipin and Liang, Li-Lin and Yang, Yu-Chen and Le, Huong Mai and Wu, Chun-Ying and others},
  journal={JMIR infodemiology},
  volume={2},
  number={1},
  pages={e37077},
  year={2022},
  publisher={JMIR Publications Inc., Toronto, Canada}
}

@article{stieglitz2013emotions,
  title={Emotions and information diffusion in social media—sentiment of microblogs and sharing behavior},
  author={Stieglitz, Stefan and Dang-Xuan, Linh},
  journal={Journal of management information systems},
  volume={29},
  number={4},
  pages={217--248},
  year={2013},
  publisher={Taylor \& Francis}
}

@inproceedings{suh2010want,
  title={Want to be retweeted? large scale analytics on factors impacting retweet in twitter network},
  author={Suh, Bongwon and Hong, Lichan and Pirolli, Peter and Chi, Ed H},
  booktitle={2010 IEEE second international conference on social computing},
  pages={177--184},
  year={2010},
  organization={IEEE}
}

@article{wang2019systematic,
  title={Systematic literature review on the spread of health-related misinformation on social media},
  author={Wang, Yuxi and McKee, Martin and Torbica, Aleksandra and Stuckler, David},
  journal={Social science \& medicine},
  volume={240},
  pages={112552},
  year={2019},
  publisher={Elsevier}
}

@article{zhang2024heart,
  title={Heart or mind? The impact of congruence on the persuasiveness of cognitive versus affective appeals in debunking messages on social media during public health crises},
  author={Zhang, Shuai and Zhang, Yang and Li, Jing and Ni, Zhenni and Liu, Zhenghao},
  journal={Computers in Human Behavior},
  volume={154},
  pages={108136},
  year={2024},
  publisher={Elsevier}
}

@article{pennycook2019fighting,
  title={Fighting misinformation on social media using crowdsourced judgments of news source quality},
  author={Pennycook, Gordon and Rand, David G},
  journal={Proceedings of the National Academy of Sciences},
  volume={116},
  number={7},
  pages={2521--2526},
  year={2019},
  publisher={National Acad Sciences}
}

@inproceedings{hanmei2013online,
  title={How online health forum users assess usergenerated content: mixed-method research},
  author={Hanmei, F and Reeva, L and Stephen, S and Shanton, C},
  booktitle={Proceedings of the 21st European Conference on Information Systems (ECIS)},
  year={2013}
}

@article{saenz2021covid,
  title={COVID-19 fake news infodemic research dataset (COVID19-FNIR dataset)},
  author={Saenz, Julio A and Gopal, Sindhu Reddy Kalathur and Shukla, Diksha},
  journal={IEEE Dataport},
  year={2021}
}

@article{egger2022topic,
  title={A topic modeling comparison between lda, nmf, top2vec, and bertopic to demystify twitter posts},
  author={Egger, Roman and Yu, Joanne},
  journal={Frontiers in sociology},
  volume={7},
  pages={886498},
  year={2022},
  publisher={Frontiers Media SA}
}

@inproceedings{cai2018interactive,
  title={Interactive Visualization for Topic Model Curation.},
  author={Cai, Guoray and Sun, Feng and Sha, Yongzhong},
  booktitle={IUI Workshops},
  year={2018}
}

@inproceedings{yu2013phrase,
  title={Phrase based topic modeling for semantic information processing in biomedicine},
  author={Yu, Zhiguo and Johnson, Todd R and Kavuluru, Ramakanth},
  booktitle={2013 12th International Conference on Machine Learning and Applications},
  volume={1},
  pages={440--445},
  year={2013},
  organization={IEEE}
}

@article{padalko2025novel,
  title={A Novel Comprehensive Framework for Detecting and Understanding Health-Related Misinformation},
  author={Padalko, Halyna and Chomko, Vasyl and Yakovlev, Sergiy and Chumachenko, Dmytro},
  journal={Information},
  volume={16},
  number={3},
  pages={175},
  year={2025},
  publisher={MDPI}
}

@article{sharifpoor2025classifying,
  title={Classifying and fact-checking health-related information about COVID-19 on Twitter/X using machine learning and deep learning models},
  author={Sharifpoor, Elham and Okhovati, Maryam and Ghazizadeh-Ahsaee, Mostafa and Avaz Beigi, Mina},
  journal={BMC Medical Informatics and Decision Making},
  volume={25},
  number={1},
  pages={73},
  year={2025},
  publisher={Springer}
}

@article{chowdhury2023understanding,
  title={Understanding misinformation infodemic during public health emergencies due to large-scale disease outbreaks: a rapid review},
  author={Chowdhury, Nashit and Khalid, Ayisha and Turin, Tanvir C},
  journal={Journal of Public Health},
  volume={31},
  number={4},
  pages={553--573},
  year={2023},
  publisher={Springer}
}

@article{cinelli2020covid,
  title={The COVID-19 social media infodemic},
  author={Cinelli, Matteo and Quattrociocchi, Walter and Galeazzi, Alessandro and Valensise, Carlo Michele and Brugnoli, Emanuele and Schmidt, Ana Lucia and Zola, Paola and Zollo, Fabiana and Scala, Antonio},
  journal={Scientific reports},
  volume={10},
  number={1},
  pages={16598},
  year={2020},
  publisher={Nature Publishing Group UK London}
}

@article{wood2018propagating,
  title={Propagating and debunking conspiracy theories on Twitter during the 2015--2016 Zika virus outbreak},
  author={Wood, Michael J},
  journal={Cyberpsychology, behavior, and social networking},
  volume={21},
  number={8},
  pages={485--490},
  year={2018},
  publisher={Mary Ann Liebert, Inc. 140 Huguenot Street, 3rd Floor New Rochelle, NY 10801 USA}
}

@article{klofstad2019drives,
  title={What drives people to believe in Zika conspiracy theories?},
  author={Klofstad, Casey A and Uscinski, Joseph E and Connolly, Jennifer M and West, Jonathan P},
  journal={Palgrave Communications},
  volume={5},
  number={1},
  pages={1--8},
  year={2019},
  publisher={Palgrave}
}

@article{tai2011rumouring,
  title={The rumouring of SARS during the 2003 epidemic in China},
  author={Tai, Zixue and Sun, Tao},
  journal={Sociology of health \& illness},
  volume={33},
  number={5},
  pages={677--693},
  year={2011},
  publisher={Wiley Online Library}
}

@article{kasereka2019cat,
  title={'The cat that kills people:’community beliefs about Ebola origins and implications for disease control in Eastern Democratic Republic of the Congo},
  author={Kasereka, Masumbuko Claude and Hawkes, Michael T},
  journal={Pathogens and global health},
  volume={113},
  number={4},
  pages={149--157},
  year={2019},
  publisher={Taylor \& Francis}
}

@article{krishnan2021research,
  title={Research note: Examining how various social media platforms have responded to COVID-19 misinformation},
  author={Krishnan, Nandita and Gu, Jiayan and Tromble, Rebekah and Abroms, Lorien C},
  journal={Harvard Kennedy School Misinformation Review},
  volume={2},
  number={6},
  pages={1--25},
  year={2021},
  publisher={Shorenstein Center for Media, Politics, and Public Policy}
}

@article{safarnejad2020contrasting,
  title={Contrasting misinformation and real-information dissemination network structures on social media during a health emergency},
  author={Safarnejad, Lida and Xu, Qian and Ge, Yaorong and Krishnan, Siddharth and Bagarvathi, Arunkumar and Chen, Shi},
  journal={American journal of public health},
  volume={110},
  number={S3},
  pages={S340--S347},
  year={2020},
  publisher={American Public Health Association}
}

@inproceedings{patwa2021fighting,
  title={Fighting an infodemic: Covid-19 fake news dataset},
  author={Patwa, Parth and Sharma, Shivam and Pykl, Srinivas and Guptha, Vineeth and Kumari, Gitanjali and Akhtar, Md Shad and Ekbal, Asif and Das, Amitava and Chakraborty, Tanmoy},
  booktitle={Combating Online Hostile Posts in Regional Languages during Emergency Situation: First International Workshop, CONSTRAINT 2021, Collocated with AAAI 2021, Virtual Event, February 8, 2021, Revised Selected Papers 1},
  pages={21--29},
  year={2021},
  organization={Springer}
}

@misc{crone2022monkeypox,
  author       = {Stephen Crone},
  title        = {Monkeypox Misinformation: Twitter Dataset},
  year         = {2022},
  howpublished = {\url{https://www.kaggle.com/datasets/stephencrone/monkeypox}},
  note         = {Accessed: 2025-07-03}
}

@article{mccarthy2010mtld,
  title={MTLD, vocd-D, and HD-D: A validation study of sophisticated approaches to lexical diversity assessment},
  author={McCarthy, Philip M and Jarvis, Scott},
  journal={Behavior research methods},
  volume={42},
  number={2},
  pages={381--392},
  year={2010},
  publisher={Springer}
}

@article{brady2017emotion,
  title={Emotion shapes the diffusion of moralized content in social networks},
  author={Brady, William J and Wills, Julian A and Jost, John T and Tucker, Joshua A and Van Bavel, Jay J},
  journal={Proceedings of the National Academy of Sciences},
  volume={114},
  number={28},
  pages={7313--7318},
  year={2017},
  publisher={National Academy of Sciences}
}

@article{amara2021multilingual,
  title={Multilingual topic modeling for tracking COVID-19 trends based on Facebook data analysis},
  author={Amara, Amina and Hadj Taieb, Mohamed Ali and Ben Aouicha, Mohamed},
  journal={Applied Intelligence},
  volume={51},
  number={5},
  pages={3052--3073},
  year={2021},
  publisher={Springer}
}

@article{yin2022sentiment,
  title={Sentiment analysis and topic modeling for COVID-19 vaccine discussions},
  author={Yin, Hui and Song, Xiangyu and Yang, Shuiqiao and Li, Jianxin},
  journal={World Wide Web},
  volume={25},
  number={3},
  pages={1067--1083},
  year={2022},
  publisher={Springer}
}

@article{vinck2019institutional,
  title={Institutional trust and misinformation in the response to the 2018--19 Ebola outbreak in North Kivu, DR Congo: a population-based survey},
  author={Vinck, Patrick and Pham, Phuong N and Bindu, Kenedy K and Bedford, Juliet and Nilles, Eric J},
  journal={The Lancet Infectious Diseases},
  volume={19},
  number={5},
  pages={529--536},
  year={2019},
  publisher={Elsevier}
}

@article{wang2019modeling,
  title={Modeling rumor propagation and mitigation across multiple social networks},
  author={Wang, Chenxu and Wang, Gaoshuai and Luo, Xiapu and Li, Hui},
  journal={Physica A: Statistical Mechanics and its Applications},
  volume={535},
  pages={122240},
  year={2019},
  publisher={Elsevier}
}

@inproceedings{Lee1999NMF,
  author    = {Daniel D. Lee and H. Sebastian Seung},
  title     = {Learning the Parts of Objects by Non-Negative Matrix Factorization},
  booktitle = {Advances in Neural Information Processing Systems},
  volume    = {12},
  pages     = {556--562},
  year      = {1999}
}

@article{zarocostas2020infodemic,
  title={How to fight an infodemic},
  author={Zarocostas, John},
  journal={The Lancet},
  volume={395},
  number={10225},
  pages={676},
  year={2020},
  publisher={Elsevier},
  doi={10.1016/S0140-6736(20)30461-X}
}

@techreport{wardle2017information,
  title={Information Disorder: Toward an Interdisciplinary Framework for Research and Policymaking},
  author={Wardle, Claire and Derakhshan, Hossein},
  institution={Council of Europe},
  year={2017},
  url={https://rm.coe.int/information-disorder-toward-an-interdisciplinary-framework-for-researc/168076277c}
}

@article{waszak2018medical,
  title={The spread of medical fake news in social media—The pilot quantitative study},
  author={Waszak, Piotr M and Kasprzycka-Waszak, Wioleta and Kubanek, Adam},
  journal={Health Policy and Technology},
  volume={7},
  number={2},
  pages={115--118},
  year={2018},
  publisher={Elsevier},
  doi={10.1016/j.hlpt.2018.03.002}
}

@article{grootendorst2022bertopic,
  title={BERTopic: Neural topic modeling with class-based TF-IDF},
  author={Grootendorst, Maarten},
  journal={arXiv preprint arXiv:2203.05794},
  year={2022}
}

@article{mcinnes2018umap,
  title={UMAP: Uniform manifold approximation and projection for dimension reduction},
  author={McInnes, Leland and Healy, John and Melville, James},
  journal={arXiv preprint arXiv:1802.03426},
  year={2018}
}

@article{campello2013density,
  title={A framework for clustering evolving data streams using HDBSCAN},
  author={Campello, Ricardo JGB and Moulavi, Davoud and Sander, J{\"o}rg},
  journal={Advances in knowledge discovery and data mining},
  pages={160--172},
  year={2013},
  publisher={Springer}
}

@article{Blei2003LDA,
  title={Latent Dirichlet Allocation},
  author={Blei, David M and Ng, Andrew Y and Jordan, Michael I},
  journal={Journal of Machine Learning Research},
  volume={3},
  pages={993--1022},
  year={2003}
}

@article{chou2020misinformation,
  author    = {Chou, Wen-Ying Sylvia and Oh, April and Klein, William M.P.},
  title     = {Addressing health-related misinformation on social media},
  journal   = {JAMA},
  volume    = {320},
  number    = {23},
  pages     = {2417--2418},
  year      = {2020}
}

@inproceedings{sikosana2024hybrid,
  author       = {Sikosana, Mkululi and Ajao, Oluwaseun and Maudsley--Barton, Sean},
  title        = {A Comparative Study of Hybrid Models in Health Misinformation Text Classification},
  booktitle    = {Proceedings of the 4th International Workshop on Open Challenges in Online Social Networks (OASIS ’24)},
  pages        = {18--25},
  address      = {Pozna{\'n}, Poland},
  month        = sep,
  year         = {2024},
  doi          = {10.1145/3677117.3685007},
  publisher    = {ACM},
  url          = {https://doi.org/10.1145/3677117.3685007}
}

@article{de2022experiments,
  title={Experiments on generalizability of BERTopic on multi-domain short text},
  author={de Groot, Muri{\"e}l and Aliannejadi, Mohammad and Haas, Marcel R},
  journal={arXiv preprint arXiv:2212.08459},
  year={2022}
}

@article{islam2020covid,
  title={COVID-19--related infodemic and its impact on public health: a global social media analysis},
  author={Islam, Md Saiful and Sarkar, Tanvir and Khan, Sazzad Hossain and Kamal, Abu-Hena Mostafa and Hasan, S M Murshid and Kabir, Abdur and Yeasmin, Dalia and Islam, Mohammad Abul and Chowdhury, Kamal Ibne Amin and Anwar, Kazi Selim and others},
  journal={The American journal of tropical medicine and hygiene},
  volume={103},
  number={4},
  pages={1621},
  year={2020},
  publisher={The American Society of Tropical Medicine and Hygiene},
  doi={10.4269/ajtmh.20-0818}
}

@article{shao2018spread,
  title={The spread of low-credibility content by social bots},
  author={Shao, Chengcheng and Ciampaglia, Giovanni Luca and Varol, Onur and Yang, Kai-Cheng and Flammini, Alessandro and Menczer, Filippo},
  journal={Nature communications},
  volume={9},
  number={1},
  pages={4787},
  year={2018},
  publisher={Nature Publishing Group UK London}
}

@article{ferrara2016rise,
  title={The rise of social bots},
  author={Ferrara, Emilio and Varol, Onur and Davis, Clayton and Menczer, Filippo and Flammini, Alessandro},
  journal={Communications of the ACM},
  volume={59},
  number={7},
  pages={96--104},
  year={2016},
  publisher={ACM New York, NY, USA}
}

@inproceedings{davis2016botornot,
  title={Botornot: A system to evaluate social bots},
  author={Davis, Clayton Allen and Varol, Onur and Ferrara, Emilio and Flammini, Alessandro and Menczer, Filippo},
  booktitle={Proceedings of the 25th international conference companion on world wide web},
  pages={273--274},
  year={2016}
}

@inproceedings{varol2017online,
  title={Online human-bot interactions: Detection, estimation, and characterization},
  author={Varol, Onur and Ferrara, Emilio and Davis, Clayton and Menczer, Filippo and Flammini, Alessandro},
  booktitle={Proceedings of the international AAAI conference on web and social media},
  volume={11},
  number={1},
  pages={280--289},
  year={2017}
}

@misc{who_mythbusters_covid,
  author       = {{World Health Organization}},
  title        = {Coronavirus disease (COVID-19) advice for the public: Mythbusters},
  year         = {2022},
  month        = jan,
  url          = {https://www.who.int/emergencies/diseases/novel-coronavirus-2019/advice-for-public/myth-busters},
  note         = {Accessed: 2026-01-02}
}

@misc{who_mpox,
  author       = {{World Health Organization}},
  title        = {Mpox},
  year         = {2024},
  month        = aug,
  url          = {https://www.who.int/news-room/fact-sheets/detail/mpox},
  note         = {Fact sheet. Published: 2024-08-26. Accessed: 2026-01-02}
}

@misc{cdc_mpox,
  author       = {{Centers for Disease Control and Prevention}},
  title        = {Monkeypox},
  year         = {2025},
  month        = sep,
  url          = {https://www.cdc.gov/monkeypox/index.html},
  note         = {Accessed: 2026-01-02}
}

@online{ukhsa_mpox,
  author  = {{UK Health Security Agency}},
  title   = {Mpox: guidance},
  date    = {2025-10},
  url     = {https://www.gov.uk/government/collections/monkeypox-guidance},
  urldate = {2026-01-02},
  note    = {Last updated: 2025-10-27}
}

@article{lee2025semi,
  title={A Semi-Automatic Labeling Framework for PCB Defects via Deep Embeddings and Density-Aware Clustering},
  author={Lee, Sang-Jeong and Seo, Sung-Bal and Bae, You-Suk},
  journal={Sensors},
  volume={25},
  number={20},
  pages={6470},
  year={2025},
  publisher={MDPI}
}

@article{meaney2023quality,
  title={Quality indices for topic model selection and evaluation: a literature review and case study},
  author={Meaney, Christopher and Stukel, Therese A and Austin, Peter C and Moineddin, Rahim and Greiver, Michelle and Escobar, Michael},
  journal={BMC Medical Informatics and Decision Making},
  volume={23},
  number={1},
  pages={132},
  year={2023},
  publisher={Springer}
}
\EOD

\end{document}